\documentclass[journal]{IEEEtran}

\usepackage{array}
\usepackage{amsthm,amsmath,amssymb}
\usepackage{multirow}
\usepackage{threeparttable}
\usepackage{booktabs}
\usepackage{enumerate}
\usepackage{cite}
\usepackage{graphicx}
\usepackage[table]{xcolor} 
\usepackage{colortbl}
\usepackage{notation}
\usepackage{subfigure}

\newcommand{\ist}{\hspace*{.3mm}}
\newcommand{\rmv}{\hspace*{-.3mm}}
\newcommand{\iist}{\hspace*{1mm}}

\newcommand{\nn}{\nonumber}

\newcommand{\trans}{\mathrm{T}}

\DeclareMathAlphabet{\mathpzc}{OT1}{pzc}{m}{it}

\begin{document}

\title{Bayesian Multiobject Tracking With Neural-Enhanced Motion and Measurement Models}

\author{Shaoxiu~Wei,~\IEEEmembership{Student Member,~IEEE,} Mingchao~Liang,~\IEEEmembership{Student Member,~IEEE,}
        and~Florian~Meyer,~\IEEEmembership{Member,~IEEE,} \vspace{-4mm}
        
                \thanks{This work was supported by the National Science Foundation (NSF) under CAREER Award No. 2146261.}

\thanks{Shaoxiu~Wei and Mingchao~Liang are with the Department of Electrical and Computer Engineering, University of California San Diego, La Jolla, CA, USA (e-mail: \texttt{shwei@ucsd.edu,m3liang@ucsd.edu}).}

\thanks{Florian~Meyer is with the Scripps Institution of Oceanography and the Department of Electrical and Computer Engineering, University of California San Diego, La Jolla, CA, USA (e-mail: \texttt{flmeyer@ucsd.edu}).}
        
        }

\providecommand{\bu}{\textcolor{blue}}
\providecommand{\rd}{\textcolor{red}}
\maketitle

\IEEEpeerreviewmaketitle

\begin{abstract}
	Multiobject tracking (MOT) is an important task in applications including autonomous driving, ocean sciences, and aerospace surveillance. Traditional MOT methods are model-based and combine sequential Bayesian estimation with data association and an object birth model. More recent methods are fully data-driven and rely on the training of neural networks. Both approaches offer distinct advantages in specific settings. In particular, model-based methods are generally applicable across a wide range of scenarios, whereas data-driven MOT achieves superior performance in scenarios where abundant labeled data for training is available. A natural thought is whether a general framework can integrate the two approaches. This paper introduces a hybrid method that utilizes neural networks to enhance aspects of the statistical motion and measurement models in Bayesian MOT that have been identified as overly simplistic. To ensure tractable computation, our framework uses belief propagation (BP) to avoid high-dimensional operations, combined with sequential Monte Carlo and sigma point (SP) techniques to perform low-dimensional operations efficiently. The resulting MOT method combines the flexibility and robustness of model-based approaches with the capability to learn complex information from data of neural networks. We evaluate the performance of the proposed method based on the nuScenes autonomous driving dataset and demonstrate that it has state-of-the-art performance. 
\end{abstract}
\begin{IEEEkeywords}
	Multiobject tracking, Bayesian framework, neural networks, LiDAR, autonomous driving.
\end{IEEEkeywords}

\section{Introduction}
\label{sec:introduction}

Multiobject tracking (MOT) \cite{BarWilTia:B11-MOT, Bla:B86-RadarTracking, BarFor:B98-JPDA,MeyThoWil:J18-BP,Mah:B14-RFS} is an important capability in a variety of applications, including autonomous driving, ocean sciences, and aerospace surveillance. Most MOT methods follow the ``detect-then-track'' paradigm. Here, at each discrete time step, a measurement detector is applied to the raw sensor data. This detector provides the measurements for Bayesian MOT. Measurements are a representation of the position, motion, and shape of objects, e.g, they can be ranges and angles, range rate, positions in Cartesian coordinates, or bounding boxes. Applying a measurement detector to the raw sensor data reduces the data rate and facilitates the development of the measurement model. However, it typically also leads to a loss of object-related information and thus to a potential reduction of tracking performance.

\par In MOT, the number of objects to be tracked is unknown, and there is a measurement-origin uncertainty, i.e., it is unknown which object generated which measurements. Thus, in addition to sequential Bayesian estimation, key aspects of MOT methods also include data association and object track management. Existing work on MOT can be divided into two types, i.e., model-based \cite{BarWilTia:B11-MOT, Bla:B86-RadarTracking, BarFor:B98-JPDA,Mah:B14-RFS,MeyThoWil:J18-BP, VoMa:J06-PHD, ReuVoVo:J14-LMB, Wil:J15-PMB} and data-driven \cite{MilRezDic:C17-RNNMOT, GuiLea:C20-GNNMOT} methods. 
\begin{figure}[!t]
	\centering
	\includegraphics[width=2.8in]{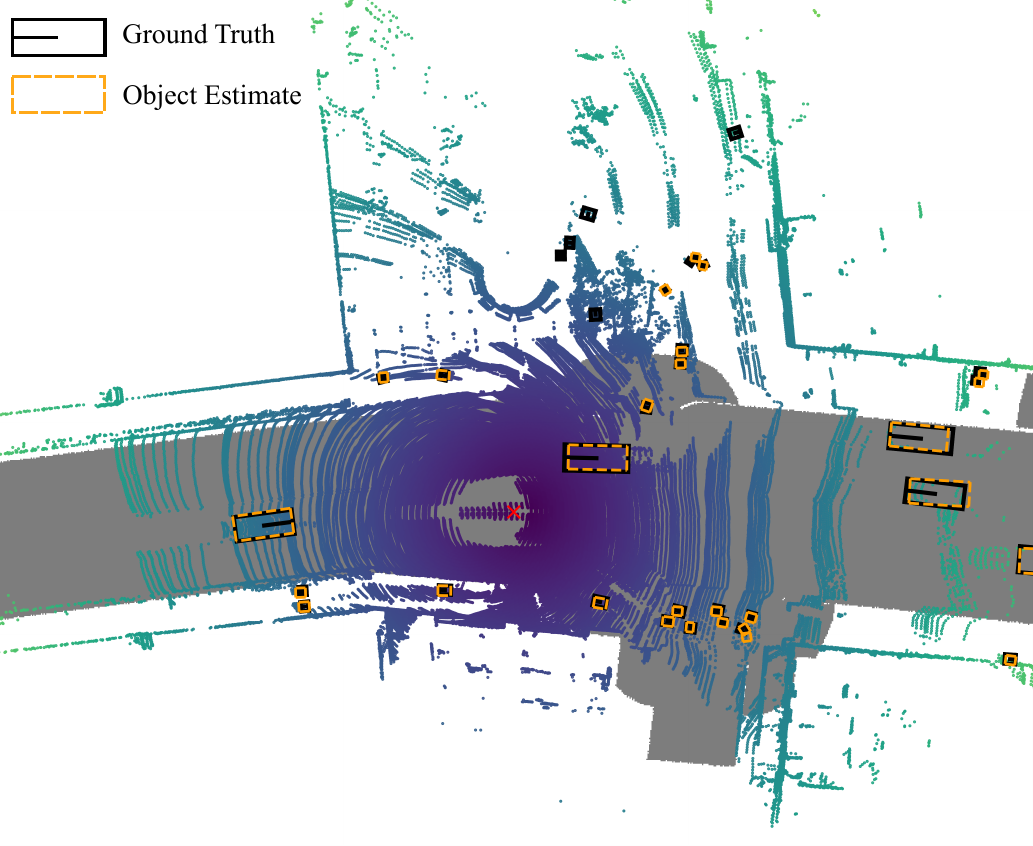}
	\caption{Autonomous driving scenario used for performance evaluation. LiDAR measurements, ground truth object states, and object state estimates provided by the proposed method are shown. The orange dashed rectangles indicate the object state estimates and the black rectangles indicate ground truth object states.}
	\label{fig:Nuscenes}
	\vspace{-4mm}
\end{figure}
While both can achieve satisfactory results in various MOT scenarios, each has potential limitations. 
\begin{itemize}
\item Model-based methods face significant challenges using complex shape feature information of objects, e.g., as provided by LiDAR or camera measurements. In particular, it turns out to be challenging to describe this type of feature information using rather simple statistical models. Moreover, to maintain tractable modeling and estimation, model-based approaches typically assume that objects move independently according to Markovian state-transition models. As a result, explicit interactions in object motion--such as those induced by parallel lanes or intersections in autonomous driving scenarios--are commonly ignored.\vspace{1mm}

\item Data-driven MOT methods typically ignore expert knowledge or any physical description of the underlying processes representing object motion and sensing. While this ``a priori information'' effectively constrains the problem, it can also provide a high degree of interpretability. Its absence leads to data-driven MOT methods being less interpretable. Furthermore, most data-driven MOT methods cannot provide any uncertainty quantification of estimation results, which may be limiting in real-world applications. 
\end{itemize}

\begin{figure*}[t]
	\centering
	\includegraphics[width=6in]{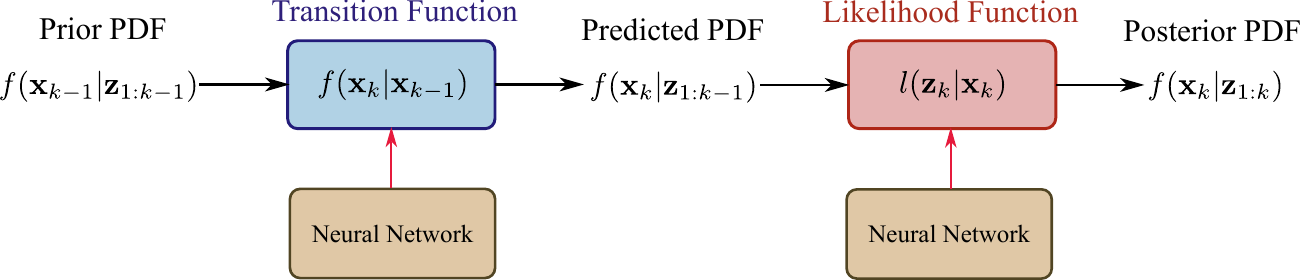}
	\caption{Block diagram of a single time step of the proposed MOT method that uses a neural network to enhance statistical motion and measurement models.}
	\label{fig:overall_bayesian}
	\vspace{-4mm}
\end{figure*}

\par  In this paper, we address limitations of model-based and data-driven estimation by integrating neural networks into the statistical model used for Bayesian MOT. Using LiDAR and camera data from an autonomous driving scenario, we demonstrate that the resulting hybrid Bayesian method can improve the robustness and overall performance of MOT. The scenario used for performance evaluation, as well as a high-level block diagram of our approach, are shown in Figs.~\ref{fig:Nuscenes} and \ref{fig:overall_bayesian}, respectively. The proposed method combines the strengths of well-designed and physically motivated statistical models with the ability of neural networks to extract complex feature information and complex interactions of object tracks. This combination is achieved by enhancing the established motion and measurement model of Bayesian MOT by the output of a neural network that has been trained using annotated data. Details on the neural-enhancement of motion and measurement model are provided in what \vspace{-0mm}follows.

The \textit{neural-enhanced motion model} is obtained by learning a state-transition function that captures the influence of historical object states and interactions with neighboring objects. The state transition function relies on a recurrent neural network (RNN) with hidden states and a weighting of neighbors inspired by the attention mechanism \cite{VasShaPar:C17-transformer}. The neural-enhanced motion model is non-Markovian and makes it possible to consider temporal dependencies and spatial relationships between objects in the prediction step of MOT by propagating a high-dimensional hidden state that acts as a sufficient statistic for previous states of the modeled object and objects in its vicinity. Sigma points (SPs) are used to perform the prediction step based on the neural-enhanced motion model efficiently \cite{JulUhl:04,MeyHliHla:J14}\vspace{-.5mm}.

A \textit{neural-enhanced measurement model} is obtained by enhancing the model-based likelihood ratios used for data association and measurement update. Building on our previous work in \cite{LiaMey:J24-NEBP,WeiLiaMey:C24-NEBP+}, our method provides two learnable factors referred to as ``affinity factors'' and ``false positive rejection factors''. The affinity factors complement the affinity information provided by model-based likelihood ratios. In particular, affinity factors act as a surrogate for a statistical model of object size and shapes, which has been identified as notoriously challenging to establish. The false positive rejection factor dynamically adjusts the false positive probability density function (PDF), individually for each measurement, using shape information extracted from raw sensor data. The neural network that provides these two factors can use different types of information as input, i.e., motion information provided by the prediction step, size measurements provided by the detector, and shape and appearance features learned from raw data\vspace{-.5mm}.

The overall structure of the statistical model is represented by a  \textit{factor graph} \cite{KscFreLoe:J01-SPA}, where certain factors representing motion and measurement model are enhanced by the neural network. Our method follows the belief propagation (BP) framework for MOT \cite{MeyThoWil:J18-BP} combined with sequential Monte Carlo and sigma point (SP) techniques \cite{AruMasGor:J02-ParticleFilter} for a fast and accurate computation of marginal posterior PDFs. The number of objects and their states are finally computed based on these marginal posterior PDFs. We evaluate the proposed method based on LiDAR and camera data from the nuScenes autonomous driving dataset \cite{Cae:C20-Nuscenes} and further assess its performance across various detectors and sensor modalities. In all tested configurations, our approach consistently achieves state-of-the-art performance.

\section{Related Work and Contributions}
The majority of existing detect-then-track MOT approaches can be broadly categorized into two types: model-based methods and data-driven methods.

Traditionally, \textit{model-based MOT methods} can be further divided into two types. The first type completely relies on Bayesian estimation. Here, object appearance and departure, object motion, data association, and measurement generation are all described by well-established statistical models. This type of methods are referred to as Bayesian MOT methods. Representative methods include vector-type methods such as the joint probabilistic data association (JPDA) filter\cite{BarFor:B98-JPDA}, the multiple hypothesis tracker (MHT)\cite{Bla:J04-MHT}, and BP \cite{MeyBraWilHla:J17,MeyThoWil:J18-BP,MeyWil:J21-BP-ExTracking,ZhaMey:J24, VenLeiTerWit:J24,WatStiTes:J25}, as well as set-type methods \cite{Mah:B14-RFS} such as probability hypothesis density (PHD) filters  \cite{VoMa:J06-PHD, VoCan:J07-CPHD}, the labeled multi-Bernoulli (LMB) filters \cite{ReuVoVo:J14-LMB, VoVo:J13-GLMB, OngVoVo:J22-GLMB}, and Poisson multi-Bernoulli (PMB) filters \cite{Wil:J15-PMB,MeyThoWil:J18-BP,GarWilGra:J18-PMBM}. 

To maintain computational tractability, model-based methods often adopt Gaussian mixture models or particle-based representations of PDFs \cite{Kay:B93-statistical-signal-processing, VoMa:J06-PHD, BarFor:B98-JPDA}.  In addition, to succeed in challenging real-world scenarios, Bayesian MOT methods that can adapt to changes in the statistical model by tracking additional nuisance parameters have been developed recently. In particular, methods that handle time-varying detection probabilities \cite{WeiZhaYi:J22-TPHD-PD, SolMeyBra:J19-BP-UnknownParameter}, motion models \cite{SolMeyBra:J19-BP-UnknownParameter, WeiZhangYi:C21-TPHD-Manoeuvring}, and object extents \cite{WeiGarYi:J25-ExTracking, GraBauReu:J17-ExTrack-Overview} have been introduced. Of particular interest for the proposed work is the BP approach for Bayesian MOT \cite{MeyBraWilHla:J17,MeyThoWil:J18-BP,MeyWil:J21-BP-ExTracking}, which relies on a factor graph representation of the underlying statistical model. The factor graph representation facilitates the identification of parts of the statistical model that are overly simplistic and provide an opportunity for neural enhancement \cite{LiaMey:J24-NEBP,WeiLiaMey:C24-NEBP+}. 

\par The second type of model-based MOT methods combines sequential Bayesian estimation with deterministic techniques or heuristic rules for data association and track management \cite{PanLiWan:C23-Tracker-SimpleTrack, LiXieLiu:C23-Tracker-PolyMOT, WanQiZha:24-Tracker-MCTrack,LiLiuWu:24-Tracker-FastPoly}. These methods typically use hand-designed rules or deterministic metrics, such as the Intersection-over-Union (IoU) and its variants, to compute association scores, representing the affinity of objects and measurements. Methods like the Hungarian algorithm or greedy matching are then commonly used to compute hard, i.e., non-probabilistic associations. Once associations are established, the Kalman filter\cite{Jaz:B70-KF} is typically adopted to perform sequential Bayesian estimation for the computation of marginal posterior PDFs of object states. This type of model-based MOT methods also perform track management based on heuristic rules.

A recent trend are \textit{data-driven methods} that employ neural architectures such as the multi-layer perceptrons (MLPs) \cite{SadJieAmb:J23-Tracker-Shasta}, convolutional neural networks (CNNs) \cite{SunAkhSon:J21-DeepTracker}, graph neural networks (GNNs) \cite{GuiLea:C20-GNNMOT, XinYinYun:C20-GNNMOT, MarAbh:J22-GNNMOT}, RNNs \cite{MilRezDic:C17-RNNMOT} and transformers \cite{ZhuHilEhs:J23-Transformer-MOT, MeiKirLea:C22-Tracker-TrackFormer} trained with annotated data for MOT. Due to the expressive power of these neural architectures, data-driven methods can potentially extract complex features and relationships from data that are difficult to model statistically. Although the underlying neural architectures vary, the primary objective of most data-driven methods is to extract information for data association and object track estimation. For example, \cite{GuiLea:C20-GNNMOT} uses message passing within a GNN for data association. Specifically, it considers novel time-aware update rules for messages by dissecting the GNN message aggregation into past and future GNN nodes. The work in \cite{MilRezDic:C17-RNNMOT} proposes a completely model-free RNN method which is capable of modeling data association, object state update, and track management within a unified RNN network structure. Furthermore, \cite{ZhuHilEhs:J23-Transformer-MOT} proposes a transformer-based MOT approach that effectively encodes long-term information across multiple frames using both temporal and spatial transformers. 

Further data-driven tracking methods that aim at establishing advanced motion models based on neural networks have been proposed in \cite{huang2023momam3t-new, Zheng2023motiontrack-new, LiuCae:C24-OfflineTracker}. In particular, \cite{huang2023momam3t-new} introduces a transformer module that aims at learning robust motion features from object tracks to improve data association. A neural architecture that captures both track histories and interactions with neighboring objects in images has been proposed in \cite{Zheng2023motiontrack-new}. The key feature of this approach is to perform the association of measurements with objects that, due to occlusion by obstacles, have not generated a measurement within a longer time window. This association is enabled by performing prediction across multiple discrete time steps.  Outside the MOT literature, methods that focus on data-driven prediction of object tracks have been presented recently in \cite{p2-new, Ivanovic2020MultimodalDG-new}. These types of methods typically consider prediction across multiple time steps by making use of a map or further aspects in the environment. A popular approach to model interactions with neighbors in data-driven prediction methods is the use of the attention mechanism \cite{huang2023momam3t-new, Zheng2023motiontrack-new, NiuZhoYu:21}.

A straightforward strategy for combining model-based and data-driven approaches for MOT is to use data-driven processing and model-based estimation for different very specific tasks. For example, in \cite{ChiLiAnb:C21-MOT}, a neural network is used for track management and data association, while a Kalman filter is employed for sequential state estimation. Our previously proposed neural-enhanced BP (NEBP) method \cite{LiaMey:J24-NEBP,WeiLiaMey:C24-NEBP+} directly integrates data-driven and model-based processing for probabilistic data association. The main idea of NEBP for MOT is to introduce a GNN that has the same structure as the loopy subgraph used for probabilistic data association in BP-based MOT, and to exchange BP and GNN messages between the two graphs. Since the GNN has been trained using labeled data and makes use of shape features learned from raw data, NEBP is expected to lead to an improved data association solution and thus to an improved overall MOT performance  \cite{SatWel:21-NEBP}. However, the loopy structure of NEBP complicates the training of the GNN, and the message exchange between BP and GNN also leads to a lack of theoretical guarantees for the resulting loopy message passing algorithm. In addition, NEBP does not neural-enhance the prediction step of MOT. Contrary to existing data-driven approaches that only aim at improving object state prediction \cite{huang2023momam3t-new,Zheng2023motiontrack-new,LiuCae:C24-OfflineTracker,huang2023momam3t-new,p2-new}, we employ a learnable, non-Markovian motion model with object interactions for calculating predicted PDFs within a Bayesian MOT framework. 
 
The presented neural-enhanced Bayesian method for MOT advances over NEBP for MOT by directly enhancing the underlying motion and measurement models. The outputs of the neural network and the statistical models are combined in a unified framework to improve the computation of posterior PDFs and object state estimates. It furthermore leads to a significantly improved MOT performance in real-world problems. The key contributions of this paper are summarized as follows.
\begin{itemize}
    \item We introduce a neural-enhanced measurement model for Bayesian MOT that makes use of complicated object shape information for data association.
    \vspace{1.3mm}
    
     \item We establish a neural-enhanced motion model for Bayesian MOT that captures the influence of historical object states and object interactions.
     \vspace{1.3mm}
    
    \item We apply the proposed method to an autonomous driving dataset based on LiDAR and camera data and demonstrate state-of-the-art object tracking performance.
    \vspace{-1mm}
\end{itemize}

 \section{Background on Model-Based MOT}
In this section, we provide necessary background on model-based MOT. If not indicated otherwise, we follow the conventional statistical model for MOT presented in \cite{MeyThoWil:J18-BP}. Before discussing the statistical model in details, we introduce basic operations performed by model-based MOT.

Let $\M{y}_k = \big[(\V{y}^1_k)^{\trans}, \dots, (\V{y}^{N_k}_k)^{\trans}\big]^{\trans}$ be the multiobject state and let $\M{z}_k$ be the measurements collected  at time $k$. The individual object state are denoted as $\V{y}^i_k$, $i \in \{1,\dots,N_k\}$, where the $I_k$ objects introduced before time step $k$ are indexed $i \in \{1,\dots,I_k\}$ and object introduced at time step $k$ are indexes $i \in \{I_k+1,\dots,N_k\}$. It is assumed that marginal posterior PDFs $f(\M{y}^{i}_{k-1}|\M{z}_{1:k-1})$ for $i \in \{1,\dots,I_k\}$ have been computed at the previous time step.

The \textit{prediction step} computes predicted marginal PDFs $f(\M{y}^{i}_{k}|$ $\M{z}_{1:k-1})$ based on marginal posterior PDFs $f(\M{y}^{i}_{k-1}| \M{z}_{1:k-1})$ for $i \in \{1,\dots,I_k\}$  from the previous time step by making use of the \textit{motion model}.
The \textit{measurement update step}, based on Bayes rule, is used to then obtain the marginal posterior PDFs, i.e.,
\begin{align}
	f(\V{y}^i_k|\M{z}_{1:k})\propto l(\M{z}_k|\M{y}^i_{k})f(\M{y}^i_k|\M{z}_{1:k-1}).\label{func:legacyUpdate}
\end{align}
Here $l(\M{z}_k|\M{y}^i_{k})$ is a statistical representation of the \textit{measurement model}. Due to measurement-origin uncertainty, the functions $l(\M{z}_k|\M{y}^i_{k})$ are coupled across objects. (This coupling is currently not indicated by our simplified notation.) Finally, for $i \in \{I_k+1,\dots,N_k\}$ the \textit{object initialization step} can be denoted\vspace{0mm} as
\begin{align}
	f(\V{y}^i_k|\M{z}_{1:k})\propto n(\M{y}^i_{k};\M{z}_k).\label{func:newPO}
\end{align} 
Here, $n(\M{y}^i_{k};\M{z}_k)$ represents the measurement and object birth model. Also $n(\M{y}^i_{k};\M{z}_k)$ is coupled across objects.

Finally, Bayesian state estimation for each object state $\M{y}^{i}_k$, $i \in \{1,\dots,N_k\}$ is performed by applying, e.g., the minimum-mean-square-error (MMSE). Bayesian estimation always relies on marginal PDFs $ f(\V{y}^i_k|\M{z}_{1:k})$.

The above steps are applied sequentially for each object state and each time step. As shown in Fig.~\ref{fig:overall_bayesian}, we aim to use neural networks to enhance the motion and measurement models, thereby enhancing the overall MOT performance. In what follows, we will describe statistical modeling for MOT in more detail.

\subsection{Object States and Motion Model}
\par At time $k$, there are $N_k$ potential objects (POs) with kinematic states $\V{x}^i_{k}$, $i\in\{1,2,\dots,N_k\}$, where $N_k$ is the maximum possible number of objects that have generated a measurement up to time $k$ \cite{MeyThoWil:J18-BP}. The kinematic state of an object $\V{x}^i_{k} \in \mathbb{R}^{d_{\mathrm{ki}}} $ consists of the object's position and further motion-related parameters. For example, in a 2-D MOT scenario, the kinematic state can be defined as $\V{x}^i_{k} = \big[(\V{p}^i_{k})^{\trans} \ist (\V{v}^i_{k})^{\trans}\big]^{\trans}\rmv\rmv\rmv\rmv$, where $\V{p}^i_{k} \in \mathbb{R}^2$ and $\V{v}^i_{k}\in \mathbb{R}^2$ are the 2-D position and velocity, respectively.

 The existence of each PO state is modeled by a binary random variable $r^i_{k}\in \{0, 1\}$. A PO state $\V{x}^i_{k}$ exists if and only if $r^i_{k}=1$. For simplicity, we use the notation ${\V{y}}^i_{k} = [(\V{x}^i_{k})^{\trans} \ist r^i_{k}]^{\trans}$ to indicate a PO state that has been augmented by an existence variable. The kinematic state $\V{x}^{i}_{k}$ of a nonexisting PO is obviously irrelevant. More specifically, for $r^i_{k} = 0$, all PDFs defined for augmented PO states, $f\big(\V{y}^{i}_{k}\big) = f\big(\V{x}^{i}_{k}\rmv, r^{i}_{k}\big)$, can be expressed as $f\big(\V{x}^i_{k}\rmv, r^{i}_{k} \!\rmv=\rmv 0 \big) = p\big( r^{i}_{k} \!\rmv=\rmv 0 \big) \ist f_{\mathrm{d}}\big(\V{x}^{i}_{k}\big)$ where $ p\big( r^{i}_{k} \!\rmv=\rmv 0 \big) \rmv\in\rmv [0,1]$ is the probability that the $i$th object does not exist at time $k$ and $ f_{\mathrm{d}}\big(\V{x}^{i}_{k}\big)$ is an arbitrary ``dummy PDF''. Similarly, for $r^i_{k} = 1$, PDFs defined for augmented PO states are given by $f\big(\V{x}^i_{k}\rmv, r^{i}_{k} \!\rmv=\rmv 1 \big) = p\big( r^{i}_{k} \!\rmv=\rmv 1 \big) \ist f\big(\V{x}^{i}_{k}\big)$, where $f(\V{x}^{i}_{k})$ is the PDF of kinematic state $\V{x}^{i}_{k}$. A measurement detector, applied to the raw sensor data, produces $J_k$ measurements denoted as $\M{z}_k = [({\V{z}^1_{k}})^{\trans}~({\V{z}^2_{k}})^{\trans}\cdots({\V{z}^{J_k}_{k}})^{\trans}]^{\trans}\rmv$\vspace{1mm}.

Generally, there are two types of PO states\vspace{-.5mm}:
\begin{enumerate}		
	\item \emph{Legacy POs} denoted as $\V{y}^i_{k} = \underline{\V{y}}^i_{k} = [(\underline{\V{x}}^i_{k})^{\trans}~\underline{r}^i_{k}]^{\trans}, i \in\{1,2,\dots,I_k\}$ represent objects that generated a measurement at a previous time step $k' < k$. The joint state of legacy POs is defined as $\underline{\V{y}}_{k} = [(\underline{\V{y}}^1_{k})^\trans\cdots(\underline{\V{y}}^{I_k}_{k})^\trans]^\trans\rmv\rmv$\vspace{1mm}.
	\item \emph{New POs} denoted as $\V{y}^i_{k}  = \overline{\V{y}}^j_{k} = [(\overline{\V{x}}^j_{k})^\trans~\overline{r}^j_{k}]^{\trans}\rmv\rmv$, $i \rmv=\rmv j \rmv+\rmv I_k$, $j \in\{1,2,\dots,J_k\}$ represent objects that generated a measurement at time step $k$ for the first time. Consequently, a new PO state $\overline{\V{y}}^j_{k}$  is introduced for each measurement $\V{z}^j_{k}, j \in \{1,2,\dots,J_k\}$. The joint state of new POs is defined as $\overline{\V{y}}_{k} = [(\overline{\V{y}}^1_{k})^\trans\cdots(\overline{\V{y}}^{J_k}_{k})^\trans]^\trans\rmv\rmv$.
\end{enumerate}
When the measurements of the next time step are considered, new POs will become legacy POs. Thus, the total number of POs is $N_k = I_k + J_k$ and the number of legacy POs at time $k$ is $I_k = I_{k-1}+ J_{k-1} = N_{k-1}$. The joint PO state $\V{y}_k$ at time $k$ is defined as $\V{y}_k = [(\underline{\V{y}}_{k})^\trans~(\overline{\V{y}}_{k})^\trans]^\trans$.

\par Contrary to the statistical model used in conventional MOT \cite{MeyThoWil:J18-BP}, we here consider a general motion model and state transition function that also considers the states of other objects and historical object states. This very general state transition function related to PO state $i$, is denoted as  $f(\underline{\M{y}}^i_k|{\M{y}}_{k_i:k-1})$, where $k_i$ is the time step where PO state with index $i$ was introduced and ${\M{y}}_{k_i:k-1}$ indicates all object states introduced since time step $k_i$. As will be elaborated in the next section, typically, only PO states that exist and are in proximity of PO state $i$ will influence the state transition of PO $\underline{\M{y}}^i_k$. This practically relevant case can be seen as a special case of the very general state-transition function $f(\underline{\M{y}}^i_k|{\M{y}}_{k_i:k-1})$.  The general state-transition function can next be expanded as
\begin{align}\label{func:transition_function_2}
	f(\underline{\M{y}}^i_k|{\M{y}}_{k_i:k-1}) =f(\underline{\M{x}}^i_k,\underline{r}^i_{k}|{\M{x}}^i_{k-1}, {r}^i_{k-1}, {\M{y}}^i_{k_i:k-2}, {\mathbf y}^{\sim i}_{k_i:k-1})
\end{align} 
 and ${\mathbf y}^{\sim i}_{k_i:k-1}$ denotes all PO states except the PO state with index $i$.

If the PO $i$ does not exist at time $k-1$, i.e., $r_{k-1}^i = 0$, then it cannot exist at time $k$ either. Thus, the state transition PDF can be expressed\vspace{-.5mm} as
\begin{align}\label{pred_existing}
	f(\underline{\M{x}}^i_k,\underline{r}^i_{k}|{\M{x}}^i_{k-1}, 0, {\M{y}}^i_{k_i:k-2}, {\mathbf y}^{\sim i}_{k_i:k-1}) = \left\{
	\begin{array}{lr}
		0, &\underline{r}^i_{k} = 1\phantom{.}\\
		f_{\mathrm{d}}(\underline{\V{x}}^i_{k}), &\underline{r}^i_{k} = 0.
	\end{array}
	\right.
\end{align}
If the PO $i$ exists at time $k-1$, i.e. $r_{k-1}^i = 1$, then it continues to exist with a survival probability $p_s$, i.e.\vspace{1mm},
\begin{align}
	&f(\underline{\M{x}}^i_k,\underline{r}^i_{k}|{\M{x}}^i_{k-1}, 1, {\M{y}}^i_{k_i:k-2}, {\mathbf y}^{\sim i}_{k_i:k-1}) \notag\\[2mm]
	&\hspace{10mm}=\begin{cases}
		\label{func:trans} p_s f( {\underline{\V{x}}}^i_{k}| {\M{x}}^i_{k_i:k-1}, {\mathbf y}^{\sim i}_{k_i:k-1}), &\hspace{1.5mm}  \underline{r}^i_{k} = 1 \\[1mm]
		(1 - p_s)f_{\mathrm{d}}(\underline{\V{x}}^i_{k}), &\hspace{1.5mm} \underline{r}^i_{k} = 0.
	\end{cases}\\[-5mm]
	\nn
\end{align}
\par However, due to high-order dependencies, establishing a statistical model for the state-transition PDF $f({\underline{\V{x}}}^i_{k}|{\M{x}}^i_{k_i:k-1}, {\mathbf y}^{\sim i}_{k_i:k-1})$ in \eqref{func:trans} is quite challenging. This complicates the computation of the predicted PDF. For this reason, most existing MOT methods assume that state transition is performed based on a Markovian state transition model and the state transition of object $i$ is independent of the state transition of all the other objects, i.e., the model in \eqref{func:trans} is simplified according to \cite{BarWilTia:B11-MOT, Bla:B86-RadarTracking, BarFor:B98-JPDA,Mah:B14-RFS,MeyThoWil:J18-BP, VoMa:J06-PHD, ReuVoVo:J14-LMB, Wil:J15-PMB}
\begin{align}
	&f(\underline{\M{x}}^i_k,\underline{r}^i_{k}|{\M{x}}^i_{k-1}, 1)=\begin{cases}
		p_s f( {\underline{\V{x}}}^i_{k}| {\M{x}}^i_{k-1}), &\hspace{1.5mm}  \underline{r}^i_{k} = 1 \nn \\[1mm]
		(1 - p_s)f_{\mathrm{d}}(\underline{\V{x}}^i_{k}), &\hspace{1.5mm} \underline{r}^i_{k} = 0.
	\end{cases}\\[-5mm]
	\nn
\end{align}
\noindent A major goal of our proposed MOT approach is to avoid the Markovian state-transition model and the assumption that objects move independently\vspace {-2mm}.


\subsection{Data Association and Measurement Model}
\par We consider a conventional MOT model that follows the point object tracking assumption, i.e., each object can generate at most one measurement, and each measurement can originate from at most one object. The former can be encoded by ``object-oriented'' data association vectors, i.e., $\V{a}_k = [a_k^1~a_k^2~{\cdots}~a_k^{I_k}]^{\trans}$\rmv\rmv\rmv, with $a_k^i \in \{0,1,\dots,J_k\}$. The latter can be encoded by ``measurement-oriented'' data association vectors, i.e., $\V{b}_k = [b_k^1~b_k^2~\cdots~b_k^{J_k}]^\trans$\rmv\rmv\rmv, with $b_k^j \in \{0,1,\dots,I_k\}$. The association vectors have the following relationships: (i) $a_k^i \rmv=\rmv j$ indicates that legacy PO $i$ is associated with measurement $j$, while $a_k^i = 0$ indicates that legacy PO $i$ is not associated to any measurement; (ii) $b_k^j = i$ indicates that measurement $j$ is associated with legacy PO $i$, while $b_k^j = 0$ indicates that measurement $j$ is not associated to any PO. A valid data association event, i.e., an event that satisfies both constraints, can be represented by both object-oriented and measurement-oriented association vectors. On the other hand, if both object-oriented and measurement-oriented association vectors can represent an event, it must be valid.

This insight motivates a hybrid representation of measurement origin uncertainty by both object-oriented and measurement-oriented vectors and to check whether $\V{a}_k$ and $\V{b}_k$ are consistent \cite{WilLau:J18,MeyThoWil:J18-BP}, i.e.\vspace{0mm},
\begin{align}
	\Phi(\V{a}_k,\V{b}_k) = \prod_{i=1}^{I_k} \ist \prod_{j=1}^{J_k} \ist \phi^{i,j}(a_k^i, b_k^j) \label{eq:highDimensional}\\[-7.5mm]
	\nn
\end{align}
where 
\begin{align}
	\phi^{i,j}(a_k^i, b_k^j)  =	
	\begin{cases}
		0, &a_k^i = j, b_k^j\neq i\\[.15mm]
		&b_k^j = i, a_k^i\neq j\\[.8mm]
		1, &\text{otherwise}.
	\end{cases}
	\label{eq:lowDimensional}
\end{align}
Note that the factors in \eqref{eq:lowDimensional} check local pairwise consistency of one entry in $\V{a}_k$ with one entry in $\V{b}_k$. If and only if all possible pairs of entries are consistent, we have global consistency, i.e., $ \Phi(\V{a}_k,\V{b}_k) \rmv=\rmv 1$ which means that  $\V{a}_k$ and $\V{b}_k$ represent the same valid data association event \cite{WilLau:J18,MeyThoWil:J18-BP}.

To perform the measurement update step in the presence of measurement-origin uncertainty, we need to compute likelihood ratios for both legacy and new POs. For legacy PO $i$, this likelihood ratio denoted $q(\underline{\V{x}}_k^i, \underline{r}_k^i, a_k^i; \V{z}_k)$, is defined\vspace{.5mm} as \cite{MeyThoWil:J18-BP}
\begin{align}
\label{func:lik_legacy_PO}	q(\underline{\V{x}}_k^i,1, a_k^i; \V{z}_k) =&~\left\{
	\begin{array}{lr}
		\frac{p_{\mathrm{d}} f(\V{z}^j_k|\underline{\V{x}}_k^i)}{C_{\mathrm{fp}}(\V{z}^j_k)}, &a_k^i = j \\
		1 - p_{\mathrm{d}}, &a_k^i = 0
	\end{array}
	\right.\\[1.5mm]
	q(\underline{\V{x}}_k^i,0, a_k^i; \V{z}_k) =&~1(a_k^i)
\end{align}
where $p_d$ is the detection probability and $C_{\mathrm{fp}}(\V{z}^j_k) = \mu_{\mathrm{fp}} \ist f_{\mathrm{fp}}(\V{z}^j_k)$ is the intensity of false positive measurements. Furthermore, $1(a_k^i)$ denotes the indicator function, i.e., $1(a_k^i)=1$ if $a_k^i=0$ and $0$ otherwise. If a measurement $j\in\{1,\dots,J_k\}$ is generated by a PO $i$, it is distributed according to $f(\V{z}^j_k|\underline{\V{x}}_k^i)$. This conditional PDF can also be interpreted of affinity of measurement $\V{z}^j_k$ with PO state $\underline{\V{x}}_k^i$. If the measurement $j$ is not generated by any PO, it is a false positive measurement that follows an uninformative PDF denoted $f_{\mathrm{fp}}(\V{z}^j_k)$. The number of false positive measurements is modeled by a Poisson PMF with mean $\mu_{\mathrm{fp}}$. 

Following the BP framework for MOT \cite{MeyThoWil:J18-BP}, the function $l(\M{z}_k|\underline{\M{y}}^i_{k})$ used in the measurement update step \eqref{func:legacyUpdate} can be obtained\vspace{-1.5mm} as 
\begin{equation}
\label{eq:likelihood}
l(\M{z}_k|\underline{\M{y}}^i_{k}) = \sum^{J_k}_{a_k^i = 0} q(\underline{\V{x}}_k^i,\underline{r}_k^i, a_k^i; \V{z}_k) \kappa(a_k^i)
\vspace{-.5mm}
\end{equation}
where data association functions $\kappa(a_k^i)$ are provided by loopy BP \cite{MeyThoWil:J18-BP}, making use of the hybrid data association representation and the corresponding consistency constraint \eqref{eq:highDimensional}.

For new PO $j$, the likelihood ratio denoted $v(\overline{\V{x}}_k^j, \overline{r}_k^j, b_k^j; \V{z}^j_k)$ is defined\vspace{-.8mm} as \cite{MeyThoWil:J18-BP}
\begin{align}
\label{func:lik_new_PO}	v(\overline{\V{x}}_k^j,1, b_k^j; \V{z}^j_k) =&~\left\{
	\begin{array}{lr}
		\frac{C_{\mathrm{n}}(\overline{\V{x}}_k^j)f(\V{z}^j_k|\overline{\V{x}}_k^j)}{C_{\mathrm{fp}}(\V{z}^j_k)}, &b_k^j = 0\\
		0, &b_k^j \ne 0
	\end{array}
	\right.\\[1.5mm]
	v(\overline{\V{x}}_k^j,0, b_k^j; \V{z}^j_k) =&~f_{\mathrm{d}}(\overline{\V{x}}_k^j). \\[-4mm]
	\nn
\end{align}
where $C_{\mathrm{n}}(\overline{\V{x}}_k^j) = \mu_{\mathrm{n}} f_{\mathrm{n}}(\overline{\V{x}}_k^j)$ is the intensity of newborn objects. The number of newborn objects is modeled by a Poisson PMF with mean $\mu_{\mathrm{n}}$. 
Note that a new PO can only exist if the corresponding measurement $\V{z}^j_k$ is not associated with any legacy PO, i.e., if $b_k^j = 0$. 

Following the BP framework for MOT \cite{MeyThoWil:J18-BP}, the function $n(\overline{\V{y}}^j_{k};\M{z}_k)$ used in the object initialization step \eqref{func:newPO}, can be obtained\vspace{-.2mm} as 
\begin{equation}
\label{eq:initialization}
n(\overline{\V{y}}^j_{k};\M{z}_k) = \sum^{I_k}_{b_k^j = 0} v(\overline{\V{x}}_k^j,\overline{r}_k^j, b_k^j; \V{z}^j_k)  \iota(b_k^j)
\vspace{-.5mm}
\end{equation}
where the data association functions $ \iota(b_k^j)$ are provided by loopy BP \cite{MeyThoWil:J18-BP}, making use of the hybrid data association representation and the corresponding consistency constraint \eqref{eq:highDimensional}.

A detailed derivation of likelihood ratios $q(\underline{\V{x}}_k^i, \underline{r}_k^i, a_k^i; \V{z}_k)$ and $v(\overline{\V{x}}_k^j, \overline{r}_k^j, b_k^j; \V{z}^j_k)$ and the corresponding measurement update steps is also provided in \cite{MeyThoWil:J18-BP}. In Section \ref{sec:updateStep}, we will discuss how the likelihood ratios can be neural enhanced by incorporating information learned from raw data not subject to measurement detection\vspace{-3mm}.

\subsection{Object Declaration and State Estimation}
\label{sec:objectDeclaration}
In our Bayesian setting, we declare the existence of POs and estimate their states based on the marginal existence probabilities $p(r^{i}_{k} \rmv=\rmv 1 | \V{z}_{1 : k})$ and the conditional PDF $f(\V{x}^{i}_{k} | r^{i}_{k} \rmv\rmv= 1, \V{z}_{1 : k})$. More specifically, a PO is declared to exist if $p(r^{i}_{k} \rmv=\rmv 1 | \V{z}_{1 : k})$ is above a predefined threshold  $T_{\mathrm{dec}}$, e.g., $T_{\mathrm{dec}}=0.5$.  For PO $i$ that is declared to exist, a state estimate of $\V{x}^{i}_{k}$ is then provided by the MMSE estimator \cite{Kay:B93-statistical-signal-processing}, i.e.\vspace{-1mm},
\begin{equation}
 \hat{\V{x}}^{i}_{k} \rmv=\rmv \int \V{x}^{i}_{k} \ist f(\V{x}^{i}_{k}  \ist|\ist r^{i}_{k} \rmv=\rmv 1, \V{z}_{1 : k}) \ist \mathrm{d}\V{x}^{i}_{k}. \nn
 \end{equation} 
Note that the existence probability and the conditional PDF are computed as $p(r^{i}_{k} \rmv= 1 \ist | \ist \V{z}_{1 : k}) \rmv=\rmv \int f(\V{x}^{i}_{k}, r^{i}_{k} \rmv=\rmv 1 \ist |$ $\V{z}_{1 : k}) \mathrm{d}\V{x}^{i}_{k}$ and $f(\V{x}^{i}_{k} \ist | \ist r^{i}_{k} \rmv=\rmv 1, \V{z}_{1 : k}) \rmv=\rmv f(\V{y}^{i}_{k} \ist | \ist \V{z}_{1 : k}) / p(r^{i}_{k} = 1 \ist | \ist \V{z}_{1 : k})$. 

As discussed previously, in this work we use BP \cite{MeyThoWil:J18-BP} to compute the marginal posterior PDFs $f(\V{y}^{i}_{k} \ist | \ist \V{z}_{1 : k}) \rmv= f(\V{x}^{i}_{k}, r^{i}_{k} \ist | \ist \V{z}_{1 : k})$. Note that, since a new PO is introduced for each measurement, the number of POs grows with time $k$. Thus, to limit computational complexity, POs whose approximate existence probability is below a threshold $T_{\mathrm{pru}}$ are removed from the state space\vspace{-1mm}.

\section{Prediction Step With Neural Enhanced Motion Model}

The goal of this section is to compute a predicted PDF that takes interactions with previous object states and neighboring object states into account. This requires addressing three challenges: (1) developing a non-Markovian motion model that involves neighboring object states, (2) establishing a computational framework to implement the motion model, and (3) performing a high-dimensional marginalization integral to compute the predicted PDF. To address these challenges, we design a motion model that can be naturally implemented by neural networks, and discuss how the predicted PDF is computed using SPs.

\subsection{Non-Markovian Motion Model With Object Interactions}
\label{sec:motionModel}

We aim to neural enhance the motion model by developing an approximate model of the state-transition PDF $f({\underline{\V{x}}}^i_{k}|{\M{x}}^i_{k_i:k-1}, {\mathbf y}^{\sim i}_{k_i:k-1})$ in \eqref{func:trans} using a neural network.  Let us introduce the vector ${\mathbf s}^i_{k}$ that consists of states of kinematic objects that exist and, at time $k$, are in the neighborhood, i.e., in the spatial vicinity, of the object with index $i$. For obvious reasons, the number of entries in  ${\mathbf s}^i_{k}$ is time-dependent, i.e., at different time steps, the number of objects that exist and are in the neighborhood of the object with index $i$ is different. The number of objects that exist and are in the neighborhood of the object with index $i$ is denoted as $M^i_k$. For practical reasons, we will limit this number to a maximum value $M$ in  Section~\ref{sec:neuralArchi}.

We assume that conditioned on ${\M{x}}^i_{k_i:k-1}$, and  $ {\mathbf s}^i_{k_i:k-1}$, the current kinematic state ${\underline{\V{x}}}^i_{k}$ is independent of all previous states of other objects that are not in its neighborhood or do not exist, i.e.\vspace{-2mm},
\begin{align}
&f({\underline{\V{x}}}^i_{k}|{\M{x}}^i_{k_i:k-1}, {\mathbf y}^{\sim i}_{k_i:k-1}) = f({\underline{\V{x}}}^i_{k}| {\M{x}}^i_{k_i:k-1},{\mathbf s}^i_{k_i:k-1}). \nn
\end{align}
In addition, it is assumed that there is a high-dimensional hidden state $\V{h}^i_{k-1} \rmv\rmv\in\rmv\rmv \mathbb{R}^{d_{\mathrm{h}}}$ that is given by $\V{h}^i_{k-1} = T\big( {\M{x}}^i_{k_i:k-2},{\mathbf s}^i_{k_i:k-2}\big)$ such that 
\begin{align}
f({\underline{\V{x}}}^i_{k}| {\M{x}}^i_{k_i:k-1},{\mathbf s}^i_{k_i:k-1}) = f({\underline{\V{x}}}^i_{k}|{\M{x}}^i_{k-1}, {\mathbf s}^i_{k-1}, \V{h}^i_{k-1}). \label{eq:proposedMotionModel}
\end{align}
Conditioned on ${\M{x}}^i_{k-1}$ and ${\mathbf s}^i_{k-1}$, the hidden state is a sufficient statistic of ${\M{x}}^i_{k_i:k-2}$ and ${\mathbf s}^i_{k_i:k-2}$ with respect to ${\underline{\V{x}}}^i_{k}$. Based these assumptions, the transition function in \eqref{func:transition_function_2} can be denoted as 
\begin{align}
	f(\underline{\M{y}}^i_k|{\M{y}}_{k_i:k-1}) = f( \underline{\M{x}}^i_k, \underline{r}^i_k | \M{x}^i_{k-1}, r^i_{k-1}, {\mathbf s}^i_{k-1}, \V{h}^i_{k-1}). \nn
\end{align} 
For the case $r^i_{k-1} = 0$, this state-transition function is equal to the right-hand side of \eqref{pred_existing}, while for $r^i_{k-1} = 1$ it is equal to the right-hand side of \eqref{func:trans} with $f({\underline{\V{x}}}^i_{k}|{\M{x}}^i_{k_i:k-1}, {\mathbf y}^{\sim i}_{k_i:k-1})$ replaced by $f({\underline{\V{x}}}^i_{k}|{\M{x}}^i_{k-1}, {\mathbf s}^i_{k-1}, \V{h}^i_{k-1})$.

 The motion model related to $f({\underline{\V{x}}}^i_{k}|{\M{x}}^i_{k-1}, {\mathbf s}^i_{k-1}, \V{h}^i_{k-1})$ in \eqref{eq:proposedMotionModel} is assumed\vspace{-1mm} as
\begin{align}
{\underline{\V{x}}}^i_{k} = g_1(\V{x}^i_{k-1}, {\mathbf s}^i_{k-1}, \V{h}^i_{k-1}) + \V{v}^i_{k} \nn \\[-4mm]
\nn
\end{align}
where $\V{v}^i_{k} \sim \mathcal{N}(\V{0};{\M Q})$ and $g_1(\V{x}^i_{k-1}, {\mathbf s}^i_{k-1}, \V{h}^i_{k-1}) $ is a learnable nonlinear function. Furthermore, it is assumed that the hidden state can be sequentially computed as
\begin{align}
	\label{func:rnn_hidden}\V{h}^i_k = g_2(\V{x}^i_{k-1}, {\mathbf s}^i_{k-1}, \V{h}^i_{k-1})
\end{align}
where $g_2(\V{x}^i_{k-1}, {\mathbf s}^i_{k-1}, \V{h}^i_{k-1})$ is a learnable function and $\V{h}^i_{k_i} = \V{0}$. Note that \eqref{func:rnn_hidden} corresponds to a transition function for the hidden state that can be written as $f(\V{h}^i_{k} |{\M{x}}^i_{k-1}, {\mathbf s}^i_{k-1}, \V{h}^i_{k-1})$. 

For future reference, we introduce the joint motion model $g(\V{x}^i_{k-1}, {\mathbf s}^i_{k-1}, \V{h}^i_{k-1}) \triangleq \big[g_1(\V{x}^i_{k-1}, {\mathbf s}^i_{k-1}, \V{h}^i_{k-1})^{\trans} \ist\ist g_2(\V{x}^i_{k-1},$ $\V{s}^i_{k-1}, \V{h}^i_{k-1})^{\trans}  \big]^{\trans}\rmv\rmv$.

\subsection{The Neural Architecture}
\label{sec:neuralArchi}

\par
\begin{figure}[!t]
	\centering
	\includegraphics[width=3.5in]{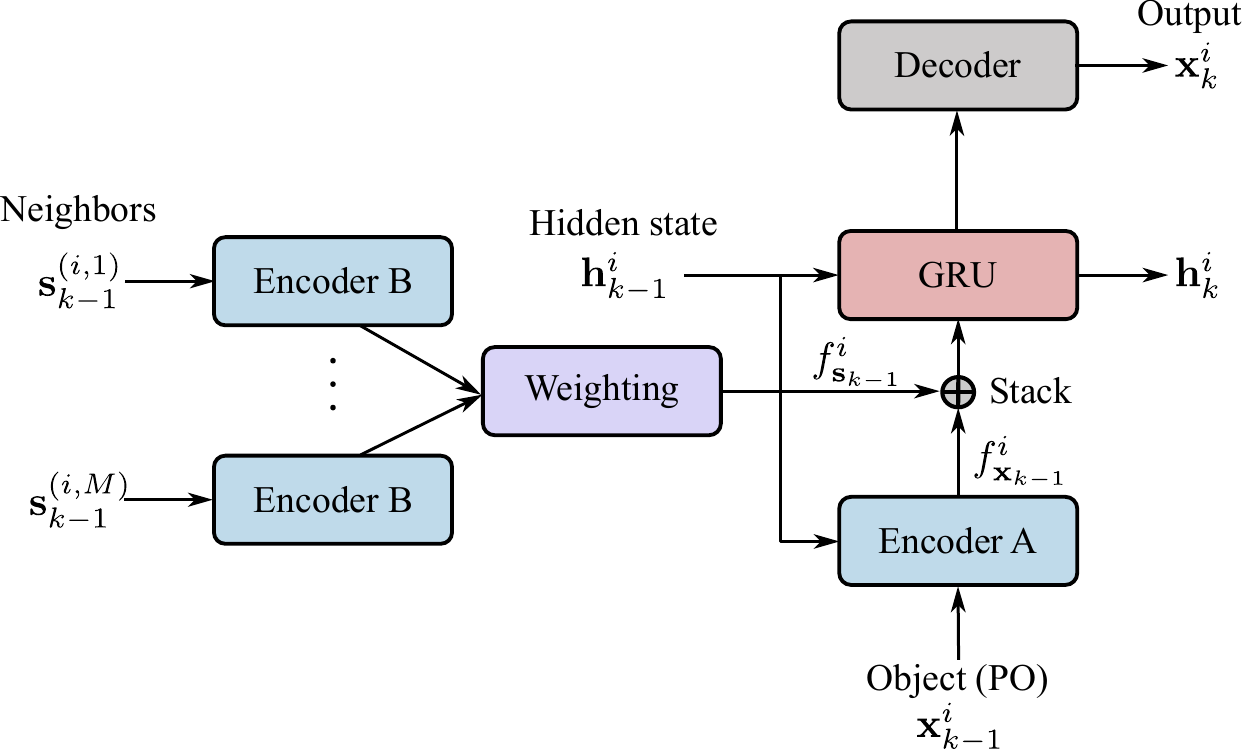}
	\vspace{-4mm}
	\caption{Proposed neural network structure as used for the neural-enhanced motion model.}
	\label{pre_net}
	\vspace{-3mm}
\end{figure}

 The neural architecture that implements the joint transition function $\big[\V{x}^i_{k} \iist \V{h}^i_{k}\big] = g(\V{x}^i_{k-1}, {\mathbf s}^i_{k-1}, \V{h}^i_{k-1})$ is discussed next. This neural architecture, shown in Fig.~\ref{pre_net}, is an extension of the RNN architecture \cite{Ale:RNN, Jun:GRU} that aims at capturing spatial interactions among objects. In particular, the hidden state $\V{h}^i_{k}$ is computed based on features extracted from the neighboring object states that are weighted based on their distance to the object of interest. This weighting of neighboring object states is inspired by the attention mechanism \cite{VasShaPar:C17-transformer}.

``Encoder A'' consists of a multilayer perceptron (MLP) with a gated recurrent unit (GRU). Based on the object state, $\V{x}^i_{k-1}$, and the hidden state, $\V{h}^i_{k-1}$, it produces a feature vector $f^i_{\V{x}_{k-1}} \rmv\in\rmv \mathbb{R}^{d_{\mathrm{h}}}$. Using the hidden state $\V{h}^i_{k-1}$ as an input to Encoder A is also an advancement over the RNN architecture that has led to significantly improved performance in our numerical evaluations. ``Encoder B'' is also a MLP. The state of neighbors indexed $m \in \{1,\dots, M\}$, i.e., each entry $s^{(i,m)}_{k-1}$, $m \rmv\in\rmv \{1,\dots,M\}$ in $ {\mathbf s}^i_{k-1}$, is processed by this encoder, resulting in feature vectors $f^{(i,m)}_{\V{s}_{k-1}} \rmv\in\rmv \mathbb{R}^{d_{\mathrm{h}}}$.

A weighting-based processing is employed to combine the feature vectors $f^{(i,m)}_{\V{s}_{k-1}}$, $m \in \{1,\dots, M\}$, i.e., to fuse information from neighboring object states. To facilitate the implementation, a maximum number of $M$ closest neighbors is considered. To measure the relevance of neighbors, distance-based weights are computed, i.e., $\mathpzc{w}_k^{(i,m)} = 1/\|{\V p}^m_k - {\V p}^i_k \|$, where $ {\V p}^i_k$ denotes the position of PO $i$ and $ {\V p}^m_k$  denotes the position of its $m$-th neighbor. Weight computation is followed by a normalization step, i.e., $w_k^{(i,m)} =\mathpzc{w}_k^{(i,m)}/\sum^M_{m=1} \mathpzc{w}_k^{(i,m)}\rmv$. The computation of weights is motivated by the fact that closer neighbors have a stronger influence on the object's motion and thus have higher relevance. 
The fused feature $f^{i}_{\V{s}_{k-1}} \rmv\in\rmv \mathbb{R}^{d_{\mathrm{h}}}$ is obtained as a weighted sum of the individual neighbors' feature vectors, i.e.\vspace{-.8mm},
\begin{equation}
	f^{i}_{\V{s}_{k-1}} = \sum^M_{m=1} { w_k^{(i,m)} f^{(i,m)}_{\V{s}_{k-1}}}. \nn
	\vspace{-.8mm}
\end{equation}
If there are fewer than $M$ neighbors, i.e., $M^i_k \rmv<\rmv M$, weights for $m \rmv>\rmv M^i_k$ are set equal to zero. We also considered a learnable attention mechanism based on Query-Key interactions  \cite{VasShaPar:C17-transformer}, but, in the autonomous driving scenario used for performance evaluation, due to limited training data, the resulting prediction performance was inferior to that of the proposed distance-based rule.

To compute an updated hidden state $\V{h}^i_{k}$, the feature vectors $f^i_{\V{x}_{k-1}}$ and $f^{i}_{\V{s}_{k-1}}$ are stacked and, together with the previous hidden state  $\V{h}^i_{k-1}$, used as an input to the GRU, i.e.\vspace{1mm},
\begin{equation}
	\V{h}^i_{k} = \mathrm{GRU}\big([f^{i \ist \trans}_{\V{x}_{k-1}} \iist f^{i \ist \trans}_{\V{s}_{k-1}}]^{\trans}, \V{h}^i_{k-1}\big).\nonumber
\end{equation}
Finally, the updated hidden state $\V{h}^i_{k}$ is processed by a decoder to obtain the updated state as
\begin{equation}
	\V{x}^i_{k} = \mathrm{Decoder}(\V{h}^i_{k}). \nn
	\vspace{-4mm}
\end{equation}

\subsection{The Neural-Enhanced Prediction Step}
\label{sec:PredictionStep}

The predicted PDF for each legacy PO is obtained via marginalization,\vspace{-1mm} i.e.,
\begin{align}
	&f(\underline{\M{x}}^i_k, \underline{r}^i_k|\M{z}_{1:k-1}) \nn\\[2mm]
	&\hspace{3mm}= \rmv\sum_{r^i_{k-1} \in \{0,1\}} \rmv\int \int \int f( \underline{\M{x}}^i_k, \underline{r}^i_k | \M{x}^i_{k-1}, r^i_{k-1}, {\mathbf s}^i_{k-1}, \V{h}^i_{k-1}) \nn\\[2mm]
	&\hspace{3mm}\times f(  \M{y}^i_{k-1}, {\mathbf s}^i_{k-1}, \V{h}^i_{k-1} |\M{z}_{1:k-1})   \mathrm{d}  \M{x}^i_{k-1}  \mathrm{d}  {\mathbf s}^i_{k-1}  \mathrm{d} \V{h}^i_{k-1}. \label{func:joint_pred}\\[-3mm]
	\nn
\end{align} 

For $\underline{r}^{i}_{k} \!\rmv=\! 0$, we have $f(\underline{\M{x}}^i_k, 0 \ist | \ist \M{z}_{1:k-1})  =  p\big( \underline{r}^{i}_{k} \!\rmv=\rmv 0 | \M{z}_{1:k-1} \big)$ $f_{\mathrm{d}}\big(\underline{\V{x}}^{i}_{k}\big)$, where $  p\big( \underline{r}^{i}_{k} \!\rmv=\rmv 0 | \M{z}_{1:k-1} \big)$ can be obtained as\vspace{.5mm} 
\begin{align}
 &p\big( \underline{r}^{i}_{k} \!\rmv=\rmv 0 | \M{z}_{1:k-1} \big) \nn\\[1.5mm]
 &\hspace{1mm}= \rmv\rmv\rmv \int \rmv\rmv f(\M{x}^i_{k-1}, 0 |\M{z}_{1:k-1}) +  \!\big(1\!-\rmv p_{\mathrm{s}} \big)  f(\M{x}^i_{k-1}, 1 |\M{z}_{1:k-1})  
\,\mathrm{d} \V{x}^{i}_{k-1} \nonumber \\[1.5mm]
&\hspace{1mm}= p\big( r^{i}_{k-1} \!\rmv=\rmv 0 |\M{z}_{1:k-1} \big) +  \!\big(1\!-\rmv p_{\mathrm{s}} \big)  p\big( r^{i}_{k-1} \!\rmv=\rmv 1 |\M{z}_{1:k-1} \big). \nn\\[-4mm]
\nn
\end{align}
Here, we used that fact for $\underline{r}^{i}_{k} \!\rmv=\rmv 0$, the state transition function in \eqref{func:joint_pred} simplifies as $f( \underline{\M{x}}^i_k, \underline{r}^{i}_{k} \!\rmv=\rmv 0 | \M{x}^i_{k-1}, r^i_{k-1}, {\mathbf s}^i_{k-1}, \V{h}^i_{k-1}) = f( \underline{\M{x}}^i_k, \underline{r}^{i}_{k} \!\rmv=\rmv 0 | \M{x}^i_{k-1}, r^i_{k-1} )$ (cf.~\eqref{pred_existing} and \eqref{func:trans}).

For $\underline{r}^i_{k} \!\rmv=\! 1$, the marginalization in \eqref{func:joint_pred} \vspace{0mm} reads
\begin{align}
&f(\underline{\M{x}}^i_k, 1 \ist | \ist \M{z}_{1:k-1})   \nn\\[1mm]
&\hspace{1mm}= p\big( r^{i}_{k-1} \!\rmv=\rmv 1 |\M{z}_{1:k-1} \big) \ist\ist p_{\mathrm{s}} \rmv \int \rmv \int \rmv \int \rmv  f({\underline{\V{x}}}^i_{k}|{\M{x}}^i_{k-1}, {\mathbf s}^i_{k-1}, \V{h}^i_{k-1}) \nn\\[1mm]
&\hspace{1mm}\times f(  {\M{x}}^i_{k-1}, {\mathbf s}^i_{k-1}, \V{h}^i_{k-1}  \ist | \ist \M{z}_{1:k-1}, r^{i}_{k-1} \!\rmv=\rmv 1)  \ist \mathrm{d} \M{x}^i_{k-1}  \mathrm{d}  {\mathbf s}^i_{k-1}  \mathrm{d}  {\mathbf h}^i_{k-1}. \nn\\[0mm]\label{eq:prediction_1Prop}\\[-7mm]
\nn
\end{align}
Similarly, we can obtain a prediction step for the hidden state as 
\begin{align}
&f(\V{h}^i_k \ist | \ist \M{z}_{1:k-1}, r^{i}_{k-1} \!\rmv=\rmv 1) \nn\\[1.5mm]
&\hspace{.5mm}= \int \int \int  f(\V{h}^i_{k} |{\M{x}}^i_{k-1}, {\mathbf s}^i_{k-1}, \V{h}^i_{k-1}) \nn\\[1mm]
&\hspace{.5mm} \times f(  {\M{x}}^i_{k-1}, {\mathbf s}^i_{k-1}, \V{h}^i_{k-1}  \ist | \ist \M{z}_{1:k-1}, r^{i}_{k-1} \!\rmv=\rmv 1)  \ist  \mathrm{d} \M{x}^i_{k-1}  \mathrm{d}  {\mathbf s}^i_{k-1}  \mathrm{d}  {\mathbf h}^i_{k-1}. \nn\\[.5mm]
\label{eq:prediction_1Proph}\\[-6.5mm]
\nn
\end{align}
Note that both $f({\underline{\V{x}}}^i_{k}|{\M{x}}^i_{k-1}, {\mathbf s}^i_{k-1}, \V{h}^i_{k-1})$ in  \eqref{eq:prediction_1Prop} and $ f(\V{h}^i_{k} |{\M{x}}^i_{k-1}, {\mathbf s}^i_{k-1}, \V{h}^i_{k-1})$ in  \eqref{eq:prediction_1Proph} are based on the joint motion model  $g(\V{x}^i_{k-1}, {\mathbf s}^i_{k-1}, \V{h}^i_{k-1})$.

To implement \eqref{eq:prediction_1Prop} and \eqref{eq:prediction_1Proph}, we furthermore approximate the marginal posterior PDF $f(  {\M{x}}^i_{k-1}, {\mathbf s}^i_{k-1}, \V{h}^i_{k-1}  \ist | \ist \M{z}_{1:k-1})$\vspace{1mm} as
\begin{align}
&f(  {\M{x}}^i_{k-1}, {\mathbf s}^i_{k-1}, \V{h}^i_{k-1}  \ist | \ist \M{z}_{1:k-1}, r^{i}_{k-1} \!\rmv=\rmv 1) \nn\\[.5mm]
&\approx f(  {\M{x}}^i_{k-1}  \ist | \ist \M{z}_{1:k-1}, r^{i}_{k-1} \!\rmv=\rmv 1) \ist f(  \V{h}^{i}_{k-1}  \ist | \ist \M{z}_{1:k-1}, r^{i}_{k-1} \!\rmv=\rmv 1) \nn\\[.5mm]
&\hspace{28mm}\times\prod^M_{m=1} f(  \V{s}^{(i,m)}_{k-1}  \ist | \ist \M{z}_{1:k-1}, r^{i}_{k-1} \!\rmv=\rmv 1) \nn\\[0mm]
&\approx \mathcal{N} \big(  \V{\mu}_{{\M{x}}^i_{k-1}}\rmv, \M{C}_{{\M{x}}^i_{k-1}} \big)  \ist \delta \big(  \V{h}^{i}_{k-1} -  \hat{\V{h}}^{i}_{k-1}  \big) \nn\\[1.5mm]
&\hspace{28mm} \times   \prod^M_{m=1} \mathcal{N}\big(  \V{\mu}_{{\M{s}}^{(i,m)}_{k-1}}\rmv, \M{C}_{{\M{s}}^{(i,m)}_{k-1}} \big),\label{eq:predApprox2}
\end{align}
where $\delta(\cdot)$ denotes the Dirac delta function. Computation of the point estimate $\hat{\V{h}}^{i}_{k-1}$ and the Gaussian parameters $\V{\mu}_{{\M{x}}^i_{k-1}}$, $\M{C}_{{\M{x}}^i_{k-1}}\rmv$, and $\V{\mu}_{{\M{s}}^{(i,m)}_{k-1}}$, $\M{C}_{{\M{s}}^{(i,m)}_{k-1}} $, $m \in \{1,\dots, M\}$ will be discussed in the last paragraph of this section. Based on the approximation in  \eqref{eq:predApprox2}, we can implement  \eqref{eq:prediction_1Prop} and \eqref{eq:prediction_1Proph} by means of SPs. An SP-based implementation is motivated by the nonlinear nature of the motion model  $g(\V{x}^i_{k-1}, {\mathbf s}^i_{k-1}, \V{h}^i_{k-1})$ and the fact that an implementation based on particles would be computationally infeasible. Note that the fact that the joint PDF $f(  {\M{x}}^i_{k-1}, {\mathbf s}^i_{k-1}, \V{h}^i_{k-1}  \ist | \ist \M{z}_{1:k-1})$ fully factorizes as in \eqref{eq:predApprox2}, is not required for an SP-based implementation. This factorization naturally arises from the computation of marginal posterior PDFs by means of BP \cite{MeyThoWil:J18-BP}.

For SP-based implementation of the high-dimensional integration in \eqref{eq:prediction_1Prop} and \eqref{eq:prediction_1Proph}, we first define a stacked mean vector $\V{\mu}^i_{k-1} \rmv\in\rmv \mathbb{R}^{L (M+1)}$ and covariance matrix $\M{C}^i_{k-1}  \rmv\in\rmv \mathbb{R}^{L (M+1) \times L (M+1)}$ corresponding to the product $\mathcal{N} \big(  \V{\mu}_{{\M{x}}^i_{k-1}}\rmv, \M{C}_{{\M{x}}^i_{k-1}} \big)   \prod^M_{m=1} \mathcal{N}\big(  \V{\mu}_{{\M{s}}^{(i,m)}_{k-1}}\rmv, \M{C}_{{\M{s}}^{(i,m)}_{k-1}} \big)$ \cite{MeyEtzLiu:J18}, i.e.\vspace{-.5mm},
\begin{align}
\V{\mu}^i_{k-1}  &\ist\triangleq\, \Big( \ist  \V{\mu}_{{\M{x}}^i_{k-1}}^{\mathrm{T}} \;\,  \V{\mu}^{\trans} _{{\M{s}}^{(i,1)}_{k-1}}\ \;\,
   \V{\mu}^{\trans} _{{\M{s}}^{(i,2)}_{k-1}} \ist\cdots\ist  \V{\mu}^{\trans} _{{\M{s}}^{(i,M)}_{k-1}}\ \Big)^{\trans} \nn\\[1mm]
\hspace{-1mm}\M{C}^i_{k-1} &\ist\triangleq\, \mathrm{bdiag} \Big\{ \M{C}_{{\M{x}}^i_{k-1}} , \M{C}_{{\M{s}}^{(i,1)}_{k-1}},
  \M{C}_{{\M{s}}^{(i,2)}_{k-1}}, \ldots,\M{C}_{{\M{s}}^{(i,M)}_{k-1}} \Big\}. \nn\\[-4mm]
\nonumber
\end{align}
where $\mathrm{bdiag}\{\cdot\}$ denotes the block diagonal matrix whose diagonal blocks are the listed matrices. This mean and block diagonal covariance matrix represent the multivariate Gaussian distribution corresponding to  \eqref{eq:predApprox2}. Next, we perform an SP-based approximation of the PDFs resulting from the integrations $\int\int\int \mathrm{d} \M{x}^i_{k-1}  \mathrm{d}  {\mathbf s}^i_{k-1}  \mathrm{d}  {\mathbf h}^i_{k-1}$ in \eqref{eq:prediction_1Prop}\ and \eqref{eq:prediction_1Proph}, respectively. A detailed motivation and implementation details for this type of SP-based approximation are provided in \cite{MeyEtzLiu:J18}.

From $\V{\mu}^i_{k-1}$ and $\M{C}^i_{k-1}$, $P \rmv=\rmv 1 \rmv+\rmv 2 L (M+1)$ weighted SPs $\big \{ \big( \omega^{(i,p)}_k\rmv, \V{x}^{(i,p)}_{k-1}, {\mathbf s}^{(i,p)}_{k-1}\big) \big\}^{P}_{p=1}$ can be computed \cite{JulUhl:04,MeyEtzLiu:J18}. SPs are next propagated through the augmented\vspace{0mm} function, i.e, $\big [\V{x}^{(i,p)}_{k}\rmv\rmv, {\mathbf h}^{(i,p)}_{k} \big] = g\big(\V{x}^{(i,p)}_{k-1}, {\mathbf s}^{(i,p)}_{k-1}, \hat{\V{h}}^i_{k-1}\big)$, $p \rmv\in\rmv \{1,\dots,P\}$\vspace{.5mm}. In this way\vspace{-.5mm}, we obtain SPs $\big \{ \big( \omega^{(i,p)}_k\rmv\rmv, \V{x}^{(i,p)}_{k} \big) \big\}^{P}_{p=1}$ and $\big \{ \big( \omega^{(i,p)}_k\rmv\rmv, {\mathbf h}^{(i,p)}_{k}\big) \big\}^{P}_{p=1}$. 

Based on these SPs, a Gaussian approximation of $f(\underline{\M{x}}^i_k, 1|\M{z}_{1:k-1} )$, is finally obtained as 
\begin{equation}
 f(\underline{\M{x}}^i_k, 1 | \V{z}_{1:k-1}) \approx p\big( r^{i}_{k-1} \!\rmv=\rmv 1 |\M{z}_{1:k-1} \big) \ist p_{\mathrm{s}} \ist\ist \mathcal{N} \big(  \V{\mu}_{{\underline{\M{x}}}^i_{k}}\rmv, \M{C}_{\underline{{\M{x}}}^i_{k}} \big).\nonumber
 \end{equation}
Here, mean $\V{\mu}_{{\underline{\M{x}}}^i_{k}}$ and the covariance matrix $ \M{C}_{\underline{{\M{x}}}^i_{k}}$ are computed\vspace{1mm} following
\vspace{-3mm}
 \begin{align}
     \V{\mu}_{{\underline{\M{x}}}^i_{k}} \rmv=&~ \sum_{p=1}^{P} \rmv \omega^{(i,p)}_k \V{x}^{(i,p)}_{k} \nn\\
     \M{C}_{\underline{{\M{x}}}^i_{k}} \rmv=&~  {\M Q} \rmv+\rmv \sum_{p=1}^{P} \rmv \omega^{(i,p)}_k \rmv (\V{x}^{(i,p)}_{k} \!-\rmv \V{\mu}_{{\underline{\M{x}}}^i_{k}})(\V{x}^{(i,p)}_{k} \!-\rmv \V{\mu}_{{\underline{\M{x}}}^i_{k}})^{\trans}. \nn \end{align}
 
 Finally, a point estimate of the hidden state  can be obtained as 
\vspace{-3mm}
 \begin{align}
     \hat{\V{h}}^{i}_{k} =&~ \sum^{P}_{p=1} \omega^{(i,p)}_{k} \ist\ist \V{h}^{(i,p)}_{k}. \nn\\[-6mm]
     \nn
 \end{align}

\section{Update Step with Neural-Enhanced Measurement Model}
\label{sec:updateStep}

In what follows, we develop the neural enhanced measurement model and the corresponding update step. The neural-enhanced measurement model involves two learnable factors referred to as ``affinity factors'' and ``false positive rejection factors'' that refine the conventional MOT measurement model. These factors are computed based on neural networks that use measurement and object-oriented features as \vspace{-2mm} input. 

\subsection{Neural-Enhanced Measurement Model}

We introduce a neural-enhanced likelihood function $f_{\mathrm{ne}}(\V{z}^j_k|\underline{\V{x}}_k^j)$. For $\V{z}^j_k$ fixed, this likelihood function is given by
\begin{equation}\label{func:affinity_factor}
	f_{\mathrm{ne}}({\V{z}}^j_k|\underline{\V{x}}^i_k) = \mathpzc{f}^{i,j}_{\mathrm{af},k} \ist f({\V{z}}^j_k|\underline{\V{x}}^i_k).
	\vspace{.5mm}
\end{equation}
In this formulation, the learnable affinity factor \(\mathpzc{f}_{\mathrm{af},k}^{i,j} > 0 \) is introduced to provide measurement-related information that is statistically independent to the model-based likelihood function $f({\V{z}}^j_k|\underline{\V{x}}^i_k)$. Specifically, $\mathpzc{f}_{\mathrm{af},k}^{i,j}$ can capture more complex feature information from measurements that is not provided by the detector or has been identified as too challenging to model statistically. It can be interpreted probabilistically as a surrogate for a statistical model describing the dependence of shape and appearance measurements on the kinematic PO state. 



In addition, the false positive rejection factor, $f^j_{\mathrm{fpr},k} \in (0,1)$, changes the false positives intensity, based on complex feature information. This factor can locally increase the false positives intensity if, based on complex features information, $\V{z}^j_k$ has been identified as potentially being a false positive measurement. The enhanced false positives intensity can be written as\vspace{-2mm}
\begin{align} \label{func:false_alarm_factor}
	C_{\mathrm{fp/ne}}(\V{z}^j_k) = C_{\mathrm{fp}}(\V{z}^j_k) / \mathpzc{f}^{j}_{\mathrm{fpr},k} =  \mu_{\mathrm{fp}}f_{\mathrm{fp}}(\V{z}^j_k) / \mathpzc{f}^{j}_{\mathrm{fpr},k} \\[-3.5mm]
	\nn
\end{align}
By plugging \eqref{func:affinity_factor} and \eqref{func:false_alarm_factor} into \eqref{func:lik_legacy_PO} and \eqref{func:lik_new_PO}, the neural-enhanced likelihood ratios are obtained by
\begin{align}
\hspace{-2mm} q_{\mathrm{ne}}(\underline{\V{x}}_k^i,1, a_k^i = j; \V{z}_k) =& ~ \mathpzc{f}^{j}_{\mathrm{fpr},k}\ist\ist \ist \mathpzc{f}^{i,j}_{\mathrm{af},k}  \ist\ist q(\underline{\V{x}}_k^i,1, a_k^i = j; \V{z}_k)\nn
\\[1.5mm]
 v_{\mathrm{ne}}(\overline{\V{x}}_k^j,1, b_k^j; \V{z}^j_k) =&~ \mathpzc{f}^{j}_{\mathrm{fpr},k}  \ist\ist v(\overline{\V{x}}_k^j,1, b_k^j; \V{z}^j_k) \nn
\end{align} 
as well as $q_{\mathrm{ne}}(\underline{\V{x}}_k^i,0, a_k^i; \V{z}_k) \rmv=\rmv q(\underline{\V{x}}_k^i,0, a_k^i; \V{z}_k)$ and $v_{\mathrm{ne}}(\overline{\V{x}}_k^i,$ $0, b_k^j; \V{z}^j_k) \rmv=\rmv v(\overline{\V{x}}_k^i,0, b_k^j; \V{z}^j_k)$. The neural enhanced likelihood ratios will replace their corresponding model-based counterpart in \eqref{eq:likelihood} and \eqref{eq:initialization}. This results in neural-enhanced functions $l_{\mathrm{ne}}(\M{z}_k|\underline{\M{y}}^i_{k})$ and $n_{\mathrm{ne}}(\M{y}^j_{k};\M{z}_k)$  used in the proposed measurement update and the new object initialization step, respectively.

\vspace{-1.5mm}.


\subsection{Feature Extraction and Computation of Learnable Factors}
\label{sec:featureExtraction}

The extraction of object and measurement-oriented features is discussed next. At time $k$ and for each legacy PO $i$, an object-oriented kinematic feature  $\underline{\V{f}}^i_{\mathrm{ki},k} \rmv\in\rmv \mathbb{R}^{d_{\mathrm{ki}}}$ is obtained as the mean of the predicted PDF of PO state $\underline{\V{x}}^{i}_{k}$, i.e.,
\begin{align}
	\underline{\V{f}}^i_{\mathrm{ki},k} = \int \underline{\V{x}}^{i}_{k} f(\underline{\V{x}}^{i}_{k} | \underline{r}^{i}_{k} = 1, \V{z}_{1 : k-1}) \ist \mathrm{d}\underline{\V{x}}^{i}_{k}. \nn
\end{align}

For shape feature extraction, the raw data is passed through the detector's ``backbone.'' The backbone provides one or multiple feature maps, a unified representation of the data on a grid that covers the region of interest for tracking. For example, in a traditional radar or sonar tracking scenario where the receiver consists of an array of antennas or hydrophones, a possible feature map is beamformed data in the range and angle domain  \cite{VanTrees:B02}. In the autonomous driving scenario considered for performance evaluation in Sections~\ref{sec:numAnal1} and \ref{sec:numAnal2}, feature maps are typically the output of a pretrained neural network and consist of a 3D tensor   \cite{YanMaoLi:J18-Voxelnet, YiZhiYu:C23-Detector-Focalformer3d, YinZhoKra:C21-Detector-Centerpoint}. The first two dimensions of the tensor represent grid points in Cartesian coordinates. For each grid point, a high-dimensional vector encodes object shape information in the region corresponding to the grid point.


\begin{figure}[t]
	\centering
	\includegraphics[width=3.5in]{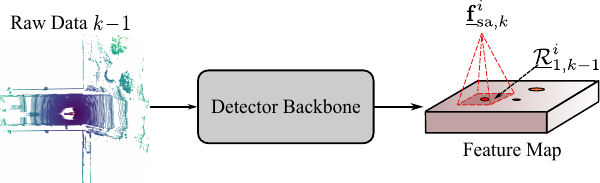}
	\caption{Extraction of an object-oriented shape feature for the case where there is $L=1$ feature map. Shape features are used in the proposed neural-enhanced measurement model\vspace*{-3mm}.}
	\label{shape_feat_extraction}
\end{figure}
\par To obtain an object-oriented shape feature vector, $\underline{\V{f}}^i_{\mathrm{sa},k}$, for legacy PO $i$ at time $k$, multiple processing stages are performed.  Fig.~\ref{shape_feat_extraction} shows these processing stages for one feature map. First, we use the kinematic information (e.g., position and rotation) from time $k\rmv-\rmv1$ as well as size information to identify a region of interest (ROI) on each feature map $l \in \{1,\dots,L\}$. Size information is either part of the kinematic state or known a priori. A ROI covers a region around the location of PO $i$ on each feature map corresponding to measurements at the last time step $k\rmv-\rmv1$. The entries of the $l$th feature map corresponding to the ROI used for further processing are denoted as $\underline{\mathcal{R}}_{l,k-1}^i$. A shape feature vector, $\underline{\V{f}}^i_{\mathrm{sa},k}$, is finally obtained by applying a neural network $\mathrm{NN}(\cdot)$ to all $\underline{\mathcal{R}}_{l,k-1}^i$, $l \rmv\in\rmv \{1,\dots,L\}$, i.e.\vspace{.5mm},
 \begin{align}
\underline{\V{f}}^i_{\mathrm{sa},k} = \big[ \mathrm{NN}(\underline{\mathcal{R}}_{1, k-1}^i)^{\trans} \iist \mathrm{NN}(\underline{\mathcal{R}}_{2,k-1}^i)^{\trans} \cdots \mathrm{NN}(\underline{\mathcal{R}}_{L,k-1}^i)^{\trans}  \big]^{\trans}\rmv\rmv\rmv\rmv. \nn
\end{align}
\par Measurement-oriented features are obtained similarly. In particular, for measurement $j$, a measurement-oriented kinematic feature, $\overline{\V{f}}^j_{\mathrm{ki},k}$, is directly provided by the output of the detector at time $k$. In addition, a measurement-oriented shape feature $\overline{\V{f}}^j_{\mathrm{sa},k}$ is extracted by (i) using feature maps corresponding to measurements from discrete time step $k$, (ii) following the same steps as for object-oriented shape features, and (iii) using the kinematic feature $\overline{\V{f}}^i_{\mathrm{ki},k}$ to identify regions of interest, $\overline{\mathcal{R}}_{l,k}^j$, on the feature maps $l \in \{1,\dots,L\}$. Typically,  neural networks used to extract object and measurement-oriented shape features share the same weights, ensuring consistency across both processes.

By using kinematic features and shape features for each PO and measurement pair, $(i,j)$, we can compute the affinity factors $\mathpzc{f}^{i,j}_{\mathrm{af},k}$ for the proposed neural-enhanced measurement model. Specifically, for each $(i,j)$, the affinity factor $\mathpzc{f}^{i,j}_{\mathrm{af},k}$  is obtained\vspace{-1mm} as	
\begin{equation}
	\mathpzc{f}^{i,j}_{\mathrm{af},k} = \mathrm{exp}\big( {\mathrm{NN}}\big(\underline{\V{f}}^i_{\mathrm{ki},k},\underline{\V{f}}^i_{\mathrm{sa},k},\overline{\V{f}}^j_{\mathrm{ki},k},\overline{\V{f}}^j_{\mathrm{sa},k} \big) \big) \nn
\end{equation}
where $\mathrm{NN}(\cdot)$ is a neural network. The exponential mapping guarantees positivity and naturally controls the factor’s dynamic range compared to the corresponding model-based likelihood function $f(\V{z}^j_k|\underline{\V{x}}_k^i)$. An optional rectified linear unit (ReLU) could be applied to $\mathpzc{f}^{i,j}_{\mathrm{af},k}$. This unit enforces that $\mathpzc{f}^{i,j}_{\mathrm{af},k} \rmv\geq\rmv 1$, i.e., the affinity factor can only increase the likelihood that measurement $j$ is associated with PO $i$ but not decrease it. It has been observed that applying the ReLU to $\mathpzc{f}^{i,j}_{\mathrm{af},k}$ can significantly reduce the training variance but potentially lead to a slightly reduced overall MOT performance. 
 The false positive rejection factor $f^j_{\mathrm{fpr},k} \in (0,1)$ for each measurement $j$, is computed by
\begin{align}
\mathpzc{f}^j_{\mathrm{fpr},k} =& \sigma\Big[{\mathrm{NN}}\big(\overline{\V{f}}^j_{\mathrm{ki},k},\overline{\V{f}}^j_{\mathrm{sa},k} \big)\Big] \nn
\end{align}
where $\sigma[\cdot]$ is the sigmoid function that restricts the output to $(0,1)$.

\section{Numerical Analysis -- Scenario and Implementation Details}
\label{sec:numAnal1}
\begin{figure*}[!t]
	\centering
	\includegraphics[scale=0.9]{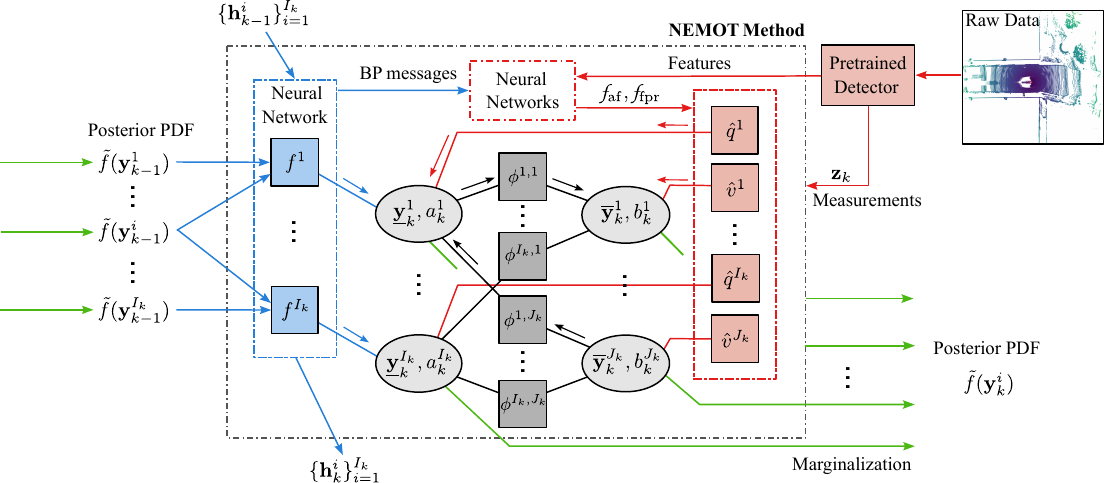}
	\caption{Block diagram of the proposed  Bayesian MOT method for a single time step $k$. The PDF computation in the proposed method is based on the indicated factor graph and corresponding message-passing procedure \cite{KscFreLoe:J01-SPA,MeyThoWil:J18-BP}. Different parts of the proposed method are indicated by different colors (1) green: input and output PDF related to time step $k$; (2) blue: computations related to the prediction step (3) red: computations related to the evaluation of the likelihood function (4) black/gray: probabilistic data association. The following shorthand notations are used: $f^i = f(\underline{\V{y}}^i_{k}|{\V{y}}^i_{k-1}, {\mathbf s}^i_{k-1}, {\mathbf h}^i_{k-1})$, $\hat q^i = \hat q(\underline{\V{y}}_k^i, a_k^i; \V{z}_k)$, $\hat v^j = \hat v(\overline{\V{y}}_k^j, b_k^j; \V{z}_k)$, $\phi^{i,j} = \phi^{i,j}(a_k^i, b_k^j)$.}
	\label{factorGraph}
\end{figure*}

In this section, we will introduce the autonomous driving scenario used for performance evaluation and discuss the loss function as well as the feature extraction mechansim. A block diagram of the proposed Bayesian MOT framework is shown in Fig.~\ref{factorGraph}\vspace{-2mm}.

\subsection{The NuScenes Autonomous Driving Dataset}

In particular, we employ LiDAR and camera data from the nuScenes autonomous driving dataset \cite{Cae:C20-Nuscenes}. This dataset consists of 1000 autonomous driving scenes and seven object classes. Tracking is performed in 2-D. The kinematic state of objects is defined as $\V{x}^i_{k} = \big[(\V{p}^i_{k})^{\trans} \ist (\V{v}^i_{k})^{\trans}\big]^{\trans}\rmv\rmv\rmv\rmv$, where $\V{p}^i_{k} \in \mathbb{R}^2$ and $\V{v}^i_{k}\in \mathbb{R}^2$ are the 2-D position and velocity, respectively. At each time step $k$, measurements are provided by a detector that consists of a neural network \cite{YiZhiYu:C23-Detector-Focalformer3d, YinZhoKra:C21-Detector-Centerpoint}. In particular, for each detected object, measurements of the 2-D position, the 2-D velocity, a reliability score, as well as the size and orientation of the bounding box are provided. The 2-D position and 2-D velocity measurement, denoted as $\V{z}^j_k \in \mathbb{R}^4$ for each $j$, is directly used in the statistical model for MOT. In particular, the likelihood 
function  $f(\V{z}^j_k|\V{x}_k^i)$ used in \eqref{func:lik_new_PO} and \eqref{func:lik_legacy_PO} is based on the linear model $\V{z}^j_k = \V{x}_k^i + \V{r}_k^j$ where $\V{r}_k^j \sim \mathcal{N}(\V{0};\M{\Sigma}_r)$ and $\M{\Sigma}_r = \mathrm{diag}\{[\sigma^2_p \ist\ist \sigma^2_p \ist\ist \sigma^2_v \ist\ist \sigma^2_v]^{\trans} \}$. Here, position and velocity-related variances $\sigma^2_p$ and $\sigma^2_v$ are determined as discussed later in this section. Size and orientation of the bounding box are used for feature extraction. The reliability score is used for final performance evaluation as also discussed later in this section.

We use the official split of the dataset, which considers using 700 scenes for training, 150 for validation, and 150 for testing. Each scene is approximately 20 seconds long and contains keyframes sampled at 2Hz. Each time step has a duration of 0.5s. We consider different detectors for the computation of measurements, mainly focusing on LiDAR detectors. If a LiDAR detector is employed, the BEV feature map used for feature extraction, as discussed in Section \ref{sec:featureExtraction}, is generated by only taking LiDAR point cloud information into account. We also consider detectors that make use of both LiDAR and camera information. Here, the BEV feature map is computed based on the recently introduced fusion technique presented in \cite{LiuTanAmi:C23-Detector-BEVFusion}, i.e., the BEV feature map combines information from both LiDAR and camera data.


\par Measurements are pre-processed using the non-maximum suppression (NMS) technique \cite{NeuVan:C06-NMS}. Here, the threshold is set to 0.1, and no score filtering is performed. As discussed in Section \ref{sec:PredictionStep}, the prediction step is implemented using SPs and results in a Gaussian representation of each PO state. The update step, on the other hand, is implemented using $10^4$ particles to represent each PO state. The Gaussian PDF resulting from the prediction step is used as a proposal PDF for the particle-based update step. From the particles resulting from the update step, a Gaussian representation is extracted by computing the sample mean and sample covariance. This Gaussian representation is then used as an input for the prediction step at the next time step. Using a particle-based implementation despite the linear sensor measurement model has been demonstrated to lead to superior data association results and thus an improved overall tracking performance \cite{KroMeyCro:J24}.

Together with the probability of existence, the reliability score is used to compute the final AMOTA for each PO. The detection region of the sensor is given by $[p^{x}_k - 54, p^{x}_k + 54] \times [p^{y}_k - 54, p^{y}_k + 54] $, where $p^{x}_k, p^{y}_k$ represent the time-dependent 2D positions of the ego vehicle. The prior PDF of false positives $f_{\mathrm{fp}}(\cdot)$ and newly detected objects $f_{\mathrm{n}}(\cdot)$ are uniformly distributed over the tracking region.The threshold for object declaration is  $T_{\mathrm{dec}}=0.5$ and the threshold for pruning is $T_{\mathrm{pru}}=10^{-4}$ (cf. Section~\ref{sec:objectDeclaration}). The survival probability is set to $p_s = 0.999$. The parameters $p_d$, $\sigma^2_p $, $\sigma^2_v$, $Q$, $C_{\mathrm{n}}(\cdot)$, $C_{\mathrm{fp}}(\cdot)$ are estimated from the training data.  In the neural-enhanced prediction step, we consider a maximum number of $M = 10$ neighbors based on their distance to the object to be predicted.

\subsection{Loss Function and Training Procedure}
\label{sec:lossFunction}

\par The computation of the loss $\ell_{k}$ at time $k$ used for the training of the proposed neural architecture can split up in a component related to the motion model, i.e.,  $\ell^{\ist \mathrm{motion}}_{k}\rmv$, and a component related to the measurement model, $\ell_k^{\mathrm{meas}}$. 

The motion-model related component of the loss is given by
\vspace{-3mm} 
\begin{align}
	\ell^{\ist \mathrm{motion}}_{k} = \frac{1}{I'_k}\sum^{I'_k}_{i=1}{|| \hat{\V{x}}_k^i - \V{x}^i_{\mathrm{gt},k}||}_1
	\label{eq:motion}
\end{align}
where we recall that $\hat{\V{x}}_k^i$ is the MMSE estimate of the kinematic state of object $i$ at time $k$ and $\V{x}^i_{\mathrm{gt},k}$ is the corresponding ground truth. For the evaluation of \eqref{eq:motion}, we need to decide which estimated object state can be associated with a ground truth state and only compute the loss for them. This decision is performed by applying the Hungarian algorithm \cite{Kuh:B55-Hungarian}. Since only some kinematic object states are associated with ground truth states, we have that $I'_k \le I_k$. 

For the measurement model-related component of the loss, we use the ``weighted'' binary cross-entropy loss \cite[Chapter 4.3]{Bis:B06}. In particular, the loss $\ell_k^{\mathrm{af}}$ for the learning of affinity factors is given by\vspace{-.3mm}
\begin{align}
\ell_{1,k}^{\mathrm{af}} =&~ - \frac{\sum_{i=1}^{I_k}\sum_{j=1}^{J_k} \mathpzc{f}_{\mathrm{gt},k}^{i,j}\ln\big(\sigma(\ln(\mathpzc{f}_{\mathrm{af},k}^{i,j})))}{\sum_{i=1}^{I_k}\sum_{j=1}^{J_k} \mathpzc{f}_{\mathrm{gt},k}^{i,j}}  \label{L_A1}\\[5.5mm]
\ell_{2,k}^{\mathrm{af}} =&~ -\frac{\sum_{i=1}^{I_k}\sum_{j=1}^{J_k}(1 - \mathpzc{f}_{\mathrm{gt},k}^{i,j})\ln\big(1 - \sigma(\ln(\mathpzc{f}_{\mathrm{af},k}^{i,j})))}{\sum_{i=1}^{I_k}\sum_{j=1}^{J_k} (1-\mathpzc{f}_{\mathrm{gt},k}^{i,j} )} \label{L_A2} \\[5.5mm]
\ell_{k}^{\mathrm{af}}  =&~ \ell_{1,k}^{\mathrm{af}} + \ell_{2,k}^{\mathrm{af}} \nn
\end{align}
where $\sigma(\cdot)$ denotes the sigmoid function and $\mathpzc{f}_{\mathrm{gt},k}^{i,j} \in \{0,1\}$ denotes the ground truth association of measurement $j$ to PO $i$, i.e., if PO $i$ is indeed matched with measurement $j$, we have $\mathpzc{f}_{\mathrm{gt},k}^{i,j}\rmv=\rmv1$, otherwise $\mathpzc{f}^{i,j}_{\mathrm{gt},k}\rmv=\rmv0$. The loss $\ell_{1,k}^{\mathrm{af}} $ characterizes accuracy of existing associations, while $\ell_{2,k}^{\mathrm{af}}$ characterizes accuracy of missing associations. The normalization terms in \eqref{L_A1} and \eqref{L_A2} address the data imbalance problem, i.e., the normalization makes sure that whether a PO $i$ is associated with a measurement $j$ or not equally contributes to the joint loss $\ell_{k}^{\mathrm{af}} $, regardless of the specific number of POs and measurements. 
\par For false positive rejection coefficients $\mathpzc{f}_{\mathrm{fpr},k}^{i,j}$, the corresponding loss function is given\vspace{1mm} by
\begin{align}
	\ell_{1,k}^{\mathrm{fpr}} &=\frac{\sum_{j=1}^{J_k} \mathpzc{f}^j_{\mathrm{gt},k}\ln(\mathpzc{f}^j_{\mathrm{fpr},k})}{\sum_{j=1}^{J_k} \mathpzc{f}^j_{\mathrm{gt},k}} \nn\\[3mm]
	\label{func:false_alarm_training} \ell_{2,k}^{\mathrm{fpr}} &= -w_{\mathrm{\mathrm{fpr}}} \ist\ist \frac{\sum_{j=1}^{J_k}(1 - \mathpzc{f}^j_{\mathrm{gt},k})\ln(1 - \mathpzc{f}^j_{\mathrm{fpr},k})}{\sum_{j=1}^{J_k} (1-\mathpzc{f}^j_{\mathrm{gt},k})}\nn\\[3mm]
	\ell_{k}^{\mathrm{fpr}}  &= \ell_{1,k}^{\mathrm{fpr}} + \ell_{2,k}^{\mathrm{fpr}} \nn\\[-4.5mm]
	\nn
\end{align}
where $\mathpzc{f}^j_{\mathrm{gt},k} \in \{0,1\}$ represents the ground truth label indicating for each measurement if it is a false positive or not, and $w_{\mathrm{\mathrm{fpr}}}\geq0$ is a parameter that makes it possible to prioritize the loss contribution $\ell_{1,k}^{\mathrm{fpr}}$ over $\ell_{2,k}^{\mathrm{fpr}}$ and vice versa. The loss contribution $\ell_{1,k}^{\mathrm{fpr}}$ represents the classification accuracy for true measurements and $\ell_{2,k}^{\mathrm{fpr}}$ represents the classification accuracy for false positives. The tuning parameter is typically $w_{\mathrm{\mathrm{fpr}}}\rmv<\rmv1$. This is motivated by the fact that missing an object is usually more harmful than generating a false positive. Finally, the measurement model-related component of the loss is obtained as $\ell^{\ist \mathrm{meas.}}_{k} = \ell_{k}^{\mathrm{af}} + \ell_{k}^{\mathrm{fpr}}$.

\par Finally, the joint loss is given by
\begin{align}
\ell =	\ell^{\ist \mathrm{motion}}_{k} + w^{\ist \mathrm{meas.}}_{\mathrm{\ell}} \ist \ell^{\ist \mathrm{meas.}}_{k} \nn
\end{align}
where $w_{\mathrm{\mathrm{fpr}}}\geq0$ is another parameter that makes it possible to prioritize the loss contribution $\ell^{\ist \mathrm{meas.}}_{k}$ over $\ell^{\ist \mathrm{motion}}_{k}$ and vice versa. This joint loss function makes it possible to train the neural enhanced motion and measurement models jointly.  
\par 
The input data for the pretraining of the motion model is generated from ground truth information. In particular, to obtain the input data, random noise is added, deterministic biases are applied, and certain states are randomly removed from tracks. For the training of the measurement model, we add additional Gaussian noise to the measurements. This is motivated by the fact that the measurements generated by detectors in the training dataset are typically of higher quality than those in the validation dataset. The joint training of the proposed method is performed based on the Adam optimizer. The batch size and learning rate are set to 1 and $10^{-4}$. The tuning parameter $w_{\mathrm{\mathrm{fpr}}} $ is 0.1. The pre-training for the transition function is based on the Adam optimizer. The batch size and learning rate are set to 32 and $10^{-3}$. The following strategy has been identified as a means to make the training more stable. First, the motion model is pretrained individually, and then its training is refined by joint training with the measurement model. For the pretraining of the motion model, we can directly use ground-truth tracks and avoid the aforementioned assignment by means of the Hungarian algorithm. 
We also explored more complex architectures --- including deeper/wider MLPs, CNN-based models, and other variants --- but observed only marginal differences in performance. Thus, we adopt a rather simple and efficient network structure with an overall depth of 10 layers.

\subsection{Feature Extraction}

The object-oriented kinematic feature is given by $\underline{\V{f}}^{i}_{\mathrm{ki},k} \rmv= [\underline{\V p}_{k}^{i \ist \trans} \ist {\underline{\V v}}^{i \ist \trans}_{k} \ist {\underline{\V u}}^{i \ist \trans}_{k}  ]^{\trans}\rmv\rmv.$ Here, $\underline{\V p}_{k}^i$ and ${\underline{\V v}}^i_{k}$ are the 2-D position and velocity estimate extracted as the mean of the predicted posterior PDF $f(  {\M{x}}^i_{k} \ist | \ist \M{z}_{1:k-1}, r^{i}_{k} \!\rmv=\rmv 1)$ and ${\underline{\V u}}^i_{k}$ represents size and orientation information.   In particular, we use bounding box information of the measurement used to initialize PO $i$ as size information. In addition, orientation information is extracted from the last measurement associated with PO $i$.  The measurement-oriented kinematic feature $\overline{\V{f}}^{j}_{\mathrm{ki},k} \rmv\rmv=\rmv\rmv [\overline{\V p}_{k}^{j \ist \trans} \ist\ist {\overline{\V v}}^{j \ist \trans}_{k} \ist\ist {\overline{\V u}}^{j \ist \trans}_{k} ]^{\trans}  $ consists of a 2-D position and velocity measurement as well as size and feature information provided by the detector.

\begin{table*}[ht]
	\centering
	\caption{Performance Evaluation Based on the Nuscenes Test Set}
	\centering
	\label{overall_sensor}
	\renewcommand{\arraystretch}{1.15} 
	\renewcommand{\arraystretch}{1.15} 
	\resizebox{\textwidth}{!}{
		\begin{tabular}{lccccccccccccc}
			\hline
			\multirow{2}{*}{\textbf{Method}} & \multirow{2}{*}{\textbf{Modalities}} & \multicolumn{8}{c}{\textbf{AMOTA}$\uparrow$} & \multirow{2}{*}{\textbf{AMOTP$\downarrow$}} & \multirow{2}{*}{\textbf{Recall$\uparrow$}} & \multirow{2}{*}{\textbf{IDS$\downarrow$}} & \multirow{2}{*}{\textbf{Frag}$\downarrow$} \\
			\cline{3-10}
			&  & \textbf{Overall} & \textbf{Bic.} & \textbf{Bus} & \textbf{Car} & \textbf{Motor} & \textbf{Ped.} & \textbf{Tra.} & \textbf{Tru.} &  &  &  &  \\
			\hline
			\rowcolor{green!25}\textbf{NEMOT}&LiDAR+Camera&77.9 &62.5 &78.4 &86.7 &82.5 &86.1 &79.3&69.8 &43.6 &0.805 &399 &259\\ 
			\hline
			MCTrack\cite{WanQiZha:24-Tracker-MCTrack} & LiDAR+Camera & 76.3 &59.1 & 77.2 &87.9 &81.9 & 83.2 & 77.9 & 66.7 &44.5 & 0.791 & 242 & 244 \\
			
			Fast-Poly\cite{LiLiuWu:24-Tracker-FastPoly} & LiDAR+Camera &75.8 & 57.3 & 76.7 & 86.3 & 82.6 & 85.2 & 76.8 & 65.6 & 47.9 & 0.776 & 326 & 270 \\
			\hline
			\rowcolor{green!25}\textbf{NEMOT}&LiDAR&75.6&51.1  &79.9 &86.5 &81.4 &85.1 &78.4 &67.1&45.6 &0.798 &353 &272\\ 
			\hline
			Poly-MOT \cite{LiXieLiu:C23-Tracker-PolyMOT} & LiDAR+Camera & 75.4 & 58.2 & 78.6 & 86.5 & 81.0 & 82.0 & 75.1 & 66.2 &  42.2 & 0.783 & 292 & 297 \\
			
			CAMO-MOT \cite{WanZhaQin:J23-CamoMOT} & LiDAR+Camera & 75.3 & 59.2 & 77.7 & 85.8 & 78.2 &85.8 & 72.3 & 67.7 & 47.2 & 0.791 & 324 & 511 \\
			
			NEBP-V3\cite{WeiLiaMey:C24-NEBP+} & LiDAR & 74.6 & 49.9 & 78.6 & 86.1 & 80.7 & 83.4 & 76.1 & 67.3 & 49.8 & 0.763 & 377 & 349 \\
			
			BEVFusion\cite{LiuTanAmi:C23-Detector-BEVFusion} & LiDAR+Camera & 74.1 & 56.0 & 77.9 & 83.1 & 74.8 & 83.7 & 73.4 & 69.5 & 40.3 & 0.779 & 506 & 422 \\
			
			MSMDFusion \cite{YiZhiYu:C23-Detector-Focalformer3d}  &LiDAR+Camera & 74.0 & 57.6 & 76.7 & 84.9 & 75.4 & 80.7 & 75.4 & 67.1 & 54.9 & 0.763 & 1088 & 743\\
			
			FocalFormer3D-F\cite{JiaJieChe:C23-Detector-MSMDFusion}   & LiDAR+Camera & 73.9 & 54.1 & 79.2 & 83.4 & 74.6 & 84.1 & 75.2 & 66.9 & 51.4 & 0.759 & 824 & 773 \\
			
			3DMOTFormer\cite{XinYinYun:C20-GNNMOT} & LiDAR+Camera & 72.5 & 48.9 & 75.4 & 83.8 & 76.0 & 83.4 & 74.3 & 65.3 & 53.9 & 0.742 & 593 & 499 \\
			
			TransFusion \cite{BaiHuZhu:C22-Detector-TransFusion} & LiDAR+Camera & 71.8 & 53.9 & 75.4 & 82.1 & 72.1 &79.6 & 73.1 & 66.3 & 55.1 & 0.758 & 944 & 673 \\
			\hline
		\end{tabular}%
	}
\end{table*}

For object-oriented shape feature extraction, position, orientation, and size information determine the ROI $\underline{\mathcal{R}}_{1,k-1}^i$. The ROI  is identified on the BEV feature map \cite{YanMaoLi:J18-Voxelnet, YiZhiYu:C23-Detector-Focalformer3d, YinZhoKra:C21-Detector-Centerpoint} in $\mathbb{R}^{180 \times 180 \times 128}$ corresponding to raw measurement from time $k-1$. Next, sample points from the ROI are determined using bilinear interpolation. In particular, we use edges or corners of the ROI. We apply two convolutional layers and a two-layer MLP to obtain the feature vector, i.e., $ \mathrm{MLP}(\mathrm{Conv}(\underline{\mathcal{R}}_{1,k-1}^i)) \rmv\in\rmv \mathbb{R}^{64}$. In addition, we perform an additional convolution over the entire BEV map and identify ROI $\underline{\mathcal{R}}_{2,k-1}^i$ on that resulting new BEV feature map in $\mathbb{R}^{180 \times 180 \times 64}$. By performing the same steps as above, we obtain a complementary feature vector  $\mathrm{MLP}(\mathrm{Conv}(\underline{\mathcal{R}}_{2,k-1}^i)) \rmv\in\rmv \mathbb{R}^{32}$. Finally, we concatenate these two feature vectors to obtain the final object-oriented shape feature, i.e.,

 \begin{align}
\hspace{-3mm}\underline{\V{f}}^i_{\mathrm{sa},k} = \big[ \mathrm{MLP}(\mathrm{Conv}(\underline{\mathcal{R}}_{1,k-1}^i))^{\trans} \iist \mathrm{MLP}(\mathrm{Conv}(\underline{\mathcal{R}}_{2,k-1}^i))^{\trans} \big]^{\trans}\rmv\rmv\rmv\rmv. \nn
\end{align}

The same feature-extraction procedure is performed to obtain measurement-oriented shape features $\overline{\V{f}}^i_{\mathrm{sa},k}$ using the BEV feature map corresponding to raw measurements from time $k$.

Based on kinematic and shape features related to a PO $i$ and measurement $j$ pair, the affinity factor $\mathpzc{f}^{i,j}_{\mathrm{af},k}$ is obtained as	
\begin{align}
	\mathpzc{f}^{i,j}_{\mathrm{af},k} =&~\mathrm{exp}(\mathrm{MLP}(\Delta {\V f}^{i,j})) \nn
\end{align}
\vspace{-2.5mm}where we\vspace{2mm} introduced
\begin{align*}
	\Delta {\V f}^{i,j}=\Big [
		\big ( {\underline {\V p}}_{k}^i - {\overline {\V p}}_{k}^j \big)^{\rmv\trans}\iist
		{\underline {\V v}}^{i \ist \trans}_{k}   \iist
		{\underline {\V u}}^{i \ist \trans}_{k}   \iist
		{\underline {\V f}}^{i \ist \trans}_{\mathrm{sa},k}  \iist
		{\overline {\V v}}^{j \ist \trans}_{k}   \iist
		{\overline {\V u}}^{j \ist \trans}_{k} \iist
		{\overline {\V f}}^{j \ist \trans}_{\mathrm{sa},k} \Big ]^{\rmv \trans}\rmv\rmv.
\end{align*}

Taking the difference of the position entries of the kinematic states removes the influence of the absolute values of position entries, which can vary strongly across scenes.
 The false positive rejection factor only relies on measurement-oriented features. In particular, for each measurement $j$, the false alarm rejection factor $\mathpzc{f}^j_{\mathrm{fpr},k}$ is obtained as 
\begin{align}
\mathpzc{f}^j_{\mathrm{fpr},k} =&~\sigma\Big[{\mathrm{MLP}}({\overline {\V v}}^j_{k},{\overline {\V u}}^j_{k},\overline{\V{f}}^j_{\mathrm{sa},k})\Big]. \nn
\end{align}

\section{Numerical Analysis -- Real-Data Results}
\label{sec:numAnal2}

In what follows, we will present results in the considered autonomous driving scenario. In particular, we will present (i) an overall performance analysis for the proposed method, (ii) an ablation study for the proposed prediction step, and (iii) an ablation study for the update step\vspace{-1mm}.

\begin{table}[!t]
	\centering
	\caption{Performance Evaluation Using Different Detectors Based on the Nuscenes Test Set}
	\label{overall_detector}
	\resizebox{\linewidth}{!}
	{\begin{tabular}{c|c|c|c|c|c} 
			\hline
			\textbf{Method}& \textbf{Detector} &\textbf{AMOTA$\uparrow$} &\textbf{AMOTP}$\downarrow$&\textbf{IDS}$\downarrow$&\textbf{Frag}$\downarrow$\\ \hline  
			\rowcolor{green!25}\textbf{NEMOT}&FocalFormer-F\cite{YiZhiYu:C23-Detector-Focalformer3d}&77.9&44.3&394&257\\ 
			\hline  
			FocalFormer3D-F\cite{YiZhiYu:C23-Detector-Focalformer3d}&FocalFormer-F&73.9&51.4&824&733\\
			\hline 
			\rowcolor{green!25}\textbf{NEMOT}&FocalFormer\cite{YiZhiYu:C23-Detector-Focalformer3d}&75.6&45.6&353&272\\ 
			\hline  
			NEBP-V3\cite{WeiLiaMey:C24-NEBP+}&FocalFormer&74.6&49.8&377&349\\ 
			FocalFormer3D\cite{YiZhiYu:C23-Detector-Focalformer3d}&FocalFormer&71.5&54.9&888
			&810\\  \hline  
			\rowcolor{green!25}\textbf{NEMOT}&Centerpoint\cite{YinZhoKra:C21-Detector-Centerpoint}&71.6&55.3&419&270\\ 
			\hline  
			ShaSTA\cite{SadJieAmb:J23-Tracker-Shasta}&Centerpoint&69.6&54.0&473
			&356\\ 
			NEBP\cite{LiaMey:J24-NEBP}&Centerpoint&68.3&62.4&227
			&299\\ 
			3DMOTFormer\cite{XinYinYun:C20-GNNMOT}&Centerpoint&68.2&49.6&438&
			529\\ 
			GNN-PMB\cite{LiuBaiXia:J23-PMB}&Centerpoint&67.8&56.0&770& 431\\ 
			ImmortalTracker\cite{WanChePan:21-ImmortalTracker}&Centerpoint&67.7 &59.9&320
			&477 \\
			Offline Tracker\cite{LiuCae:C24-OfflineTracker}&Centerpoint&67.1&52.2&570 &592\\ 
			SimpleTrack\cite{PanLiWan:C23-Tracker-SimpleTrack}&Centerpoint &66.8 &55.0&575& 591\\
			BP\cite{MeyThoWil:J18-BP}&Centerpoint&66.6&57.1&182& 245\\ 	
			\hline  
		\end{tabular}
	}
	\vspace{-1mm}
\end{table}

\subsection{Overall Performance Evaluation}
The two primary metrics for evaluating nuScenes tracking results are the Average Multi-Object Tracking Accuracy (AMOTA) and Average Multi-Object Tracking Precision (AMOTP) \cite{Cae:C20-Nuscenes}. The AMOTA includes errors such as false positives, missed objects, and identity switches, while AMOTP consists of position errors between ground truth and estimated tracks. Secondary metrics include, e.g. (i) the ``IDS'', which represents the total number of identity switches and (ii) the ``Frag'', which represents the total number of times a trajectory is fragmented (i.e., interrupted during tracking). Details about these metrics are provided in \cite{Cae:C20-Nuscenes}.

\par The performance of the proposed NEMOT method is presented in Tables \ref{overall_sensor} and \ref{overall_detector}. The ranking of the different methods is determined based on the AMOTA, which we consider the primary metric in our paper, consistent with the official NuScenes benchmark. The results demonstrate that even if combined with the ``FocalFormer''  detector \cite{YiZhiYu:C23-Detector-Focalformer3d}, the proposed NEMOT method achieves the highest AMOTA and competitive AMOTP scores, compared to all other reference methods. A comparison across different methods based on Table \ref{overall_sensor} shows that, if both LiDAR and camera measurements are considered, our method achieves the best performance for trailers, truck, bicycle and pedestrian, and ranks second place for cars and motorcycles. Bicycles are notoriously challenging to distinguish from pedestrians in LiDAR point clouds. This insight explains the noticeable performance gap in AMOTA performance between methods that only rely on LiDAR data and methods that make use of LiDAR and camera data. 

\label{sec:ablStudy}
\begin{table}[!t]
	\centering
	\caption{Performance Evaluation of Neural-Enhanced Prediction and Update Steps Based on Nuscenes Validation Set}
	\label{tab:ablation_pred_update}
		\begin{tabular}{c|c|c} 
			\hline
			\textbf{Method} &{\textbf{AMOTA}$\uparrow$} &\textbf{AMOTP}$\downarrow$\\ 
			\hline
			$\text{Prediction}(\times), \text{Update}(\times)$ &75.5&52.0\\   
			$\text{Prediction}(\times), \text{Update}(\checkmark)$&76.7&49.5\\ 
			$\text{Prediction}(\checkmark), \text{Update}(\times)$&76.6&50.0\\ 
			$\text{Prediction}(\checkmark), \text{Update}(\checkmark)$ &77.2&49.7\\ 
			\hline
	\end{tabular}
	\vspace{-3mm}
\end{table}
\par Also when combined with the ``CenterPoint'' detector \cite{YinZhoKra:C21-Detector-Centerpoint}, NEMOT outperforms all reference techniques. Additionally, as shown in Table \ref{overall_detector}, NEMOT can achieve a lower IDS and Frag metric compared to reference methods including ``ShaSTA'' \cite{SadJieAmb:J23-Tracker-Shasta}, ``3DMOTFormer'' \cite{XinYinYun:C20-GNNMOT} or ``SimpleTrack'' \cite{PanLiWan:C23-Tracker-SimpleTrack}. This is because the proposed method uses a statistical model to determine the initialization and termination of tracks, which is more robust compared to the heuristic track management performed by these reference methods.

\subsection{Ablation Study}
\begin{figure*}[!t]
	\centering
	\subfigure[Pedestrian, $k=0$]{
		\includegraphics[width=2.2in]{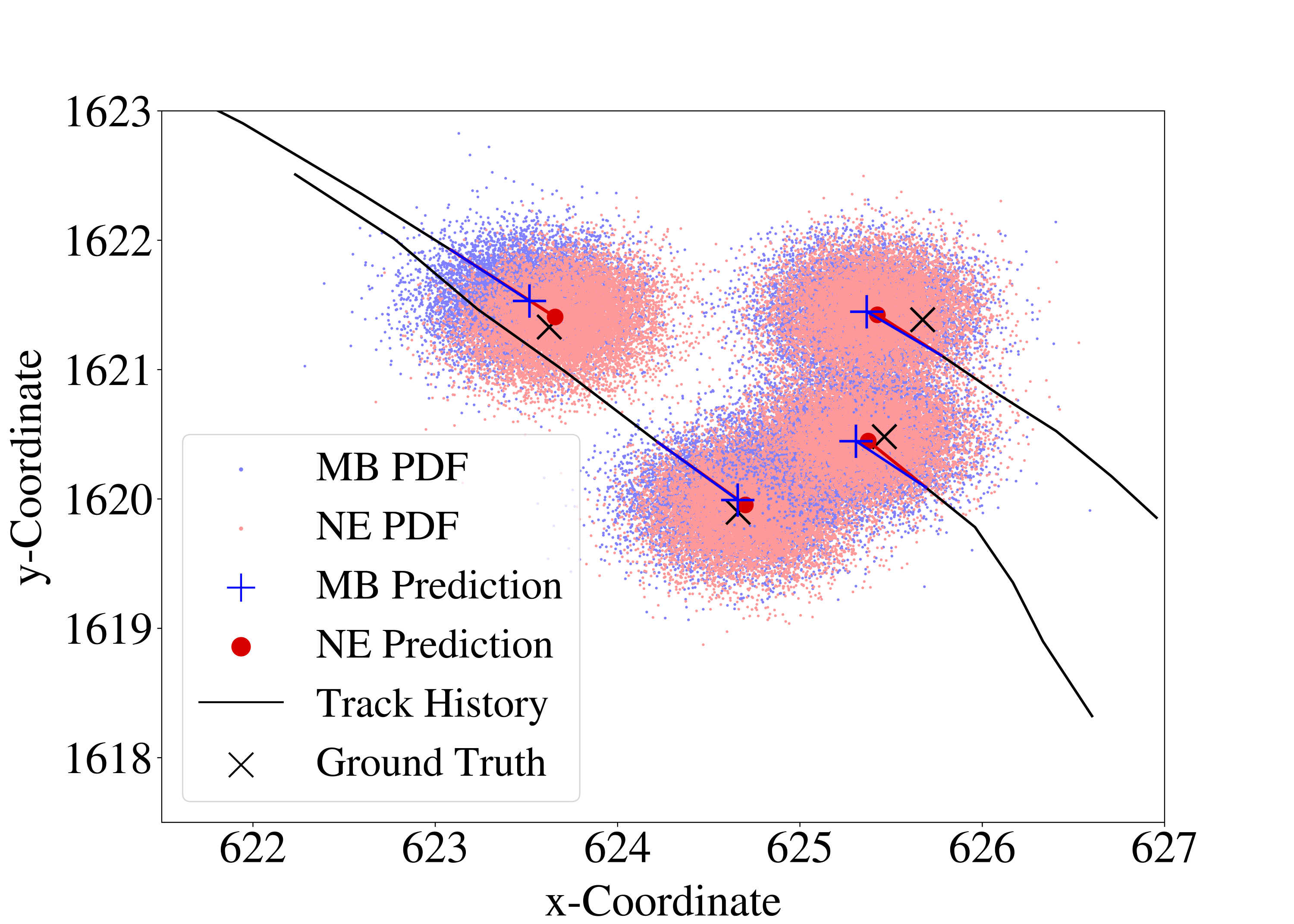}
	}
	\subfigure[Pedestrian, $k=1$]{
		\includegraphics[width=2.2in]{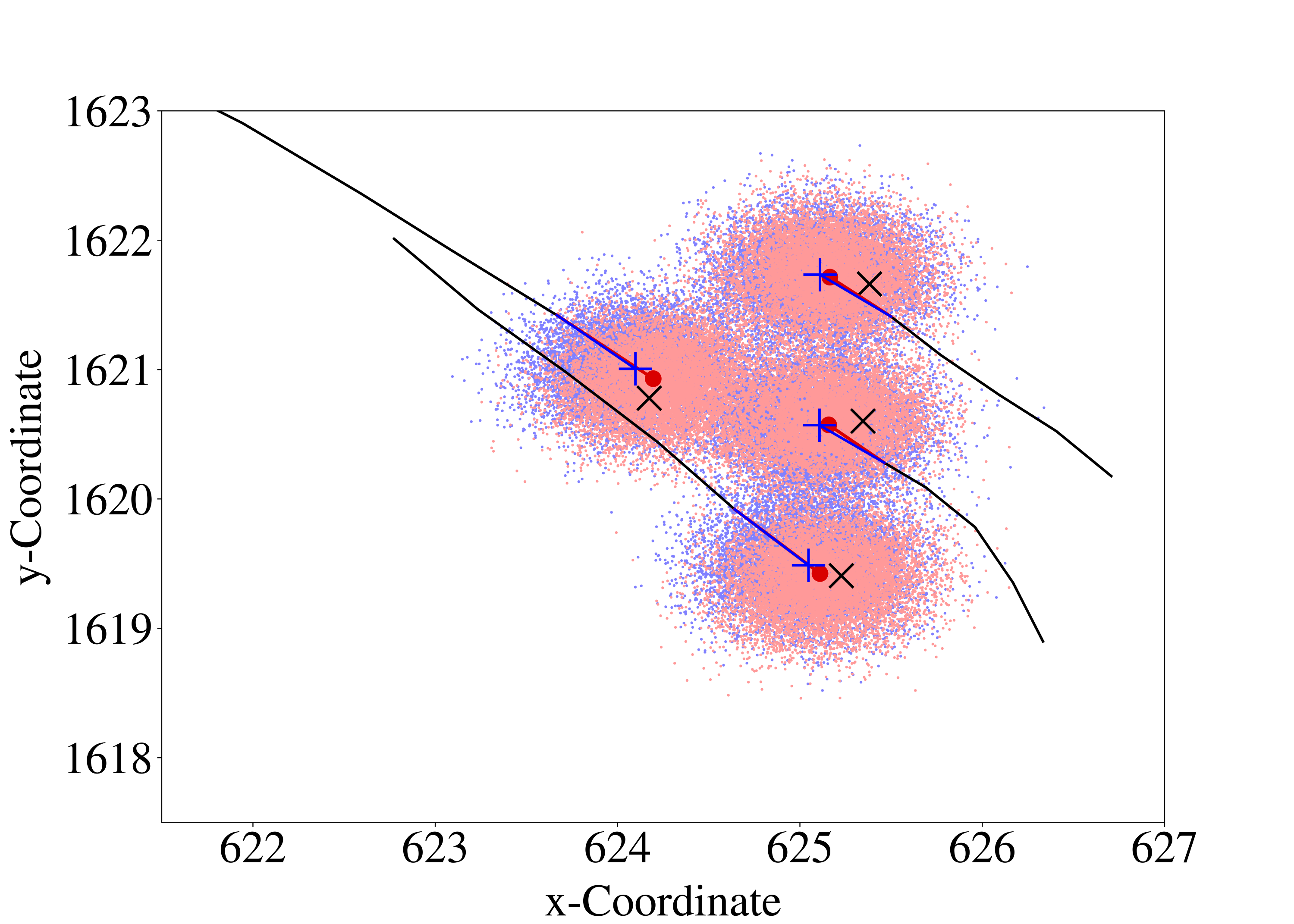}
	}
	\subfigure[Pedestrian, $k=2$]{
			\includegraphics[width=2.2in]{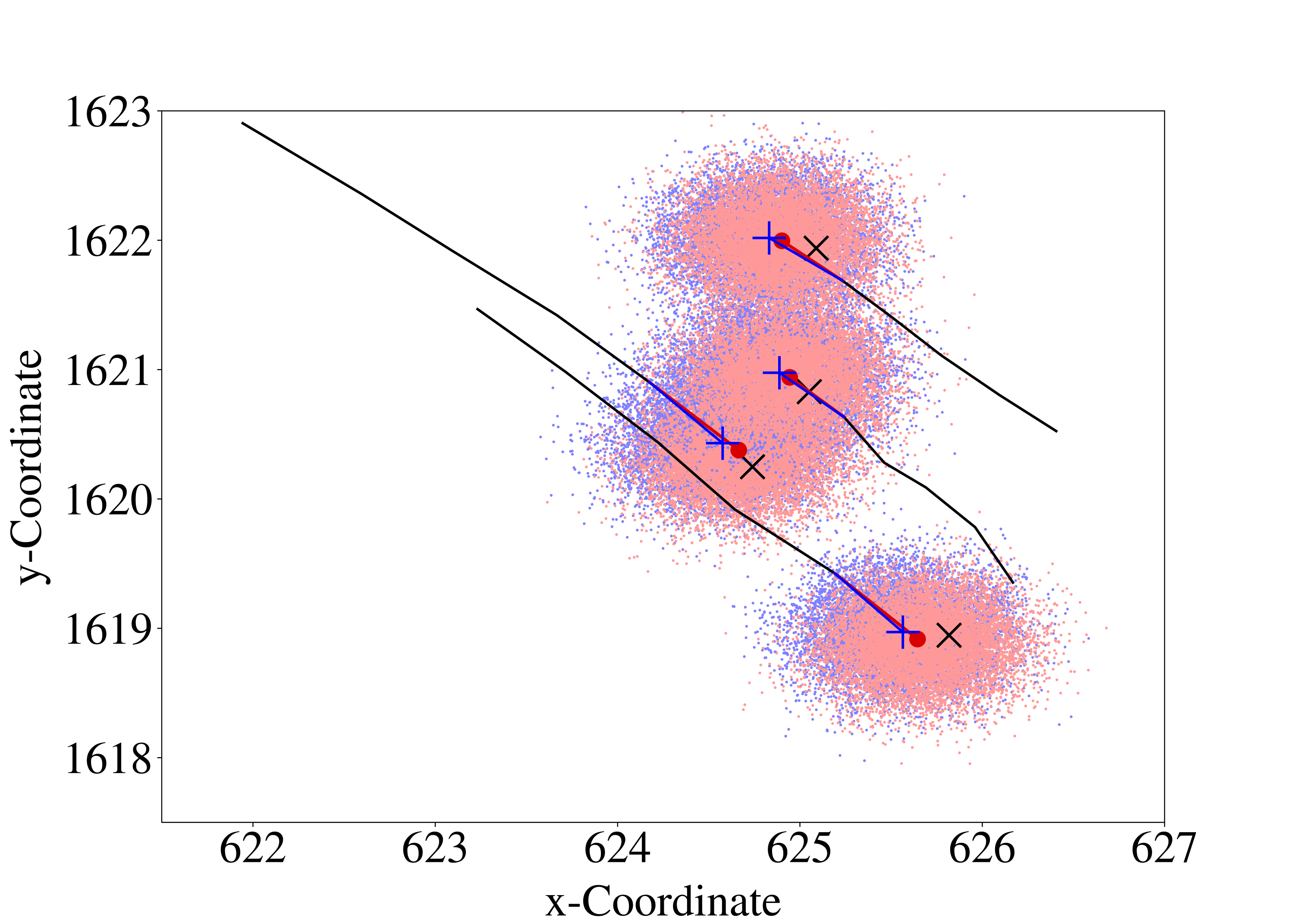}
	}
	\centering
	\subfigure[Car, $k=0$]{
		\includegraphics[width=2.2in]{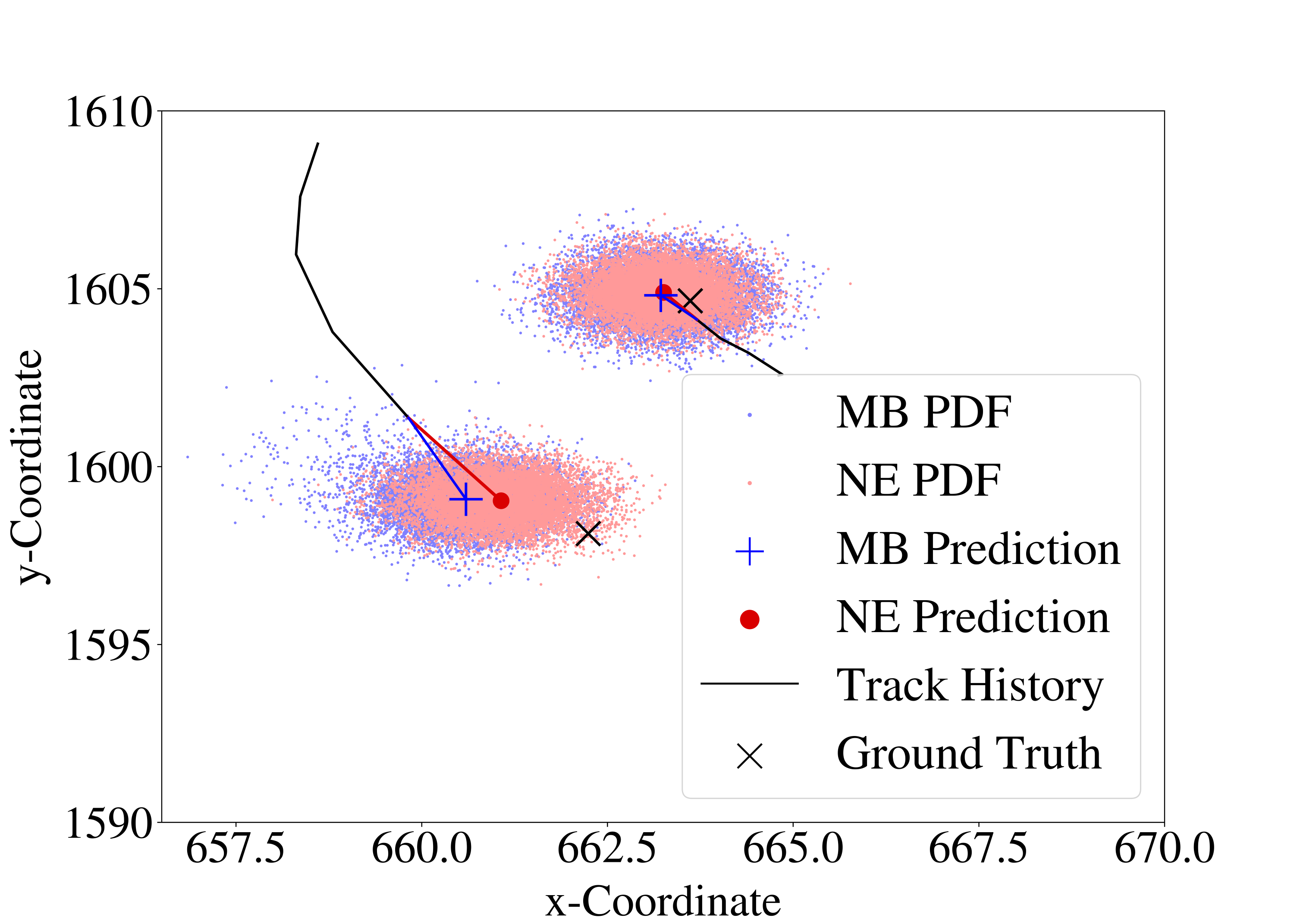}
	}
	\subfigure[Car, $k=1$]{
		\includegraphics[width=2.2in]{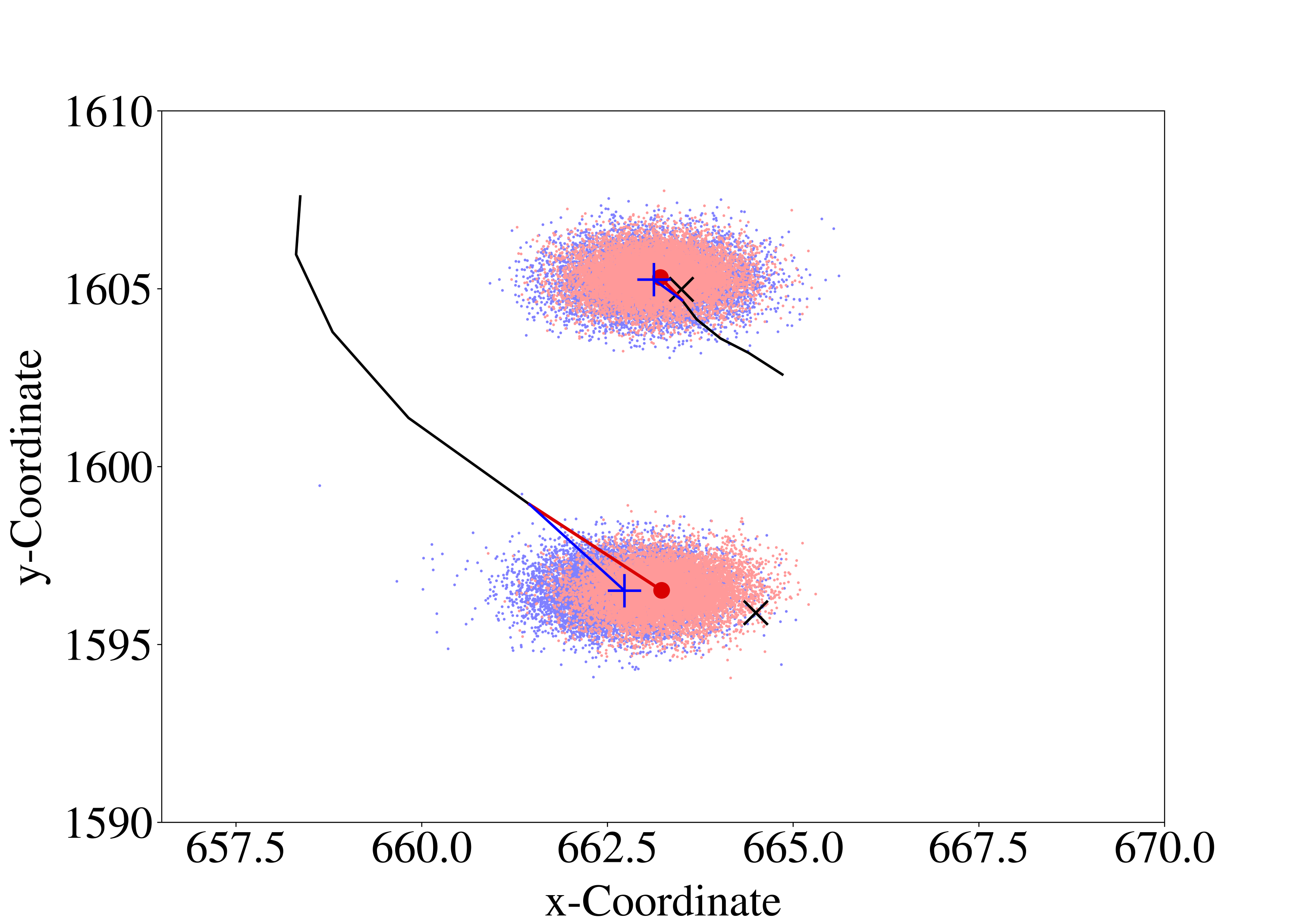}
	}
	\subfigure[Car, $k=2$]{
		\includegraphics[width=2.2in]{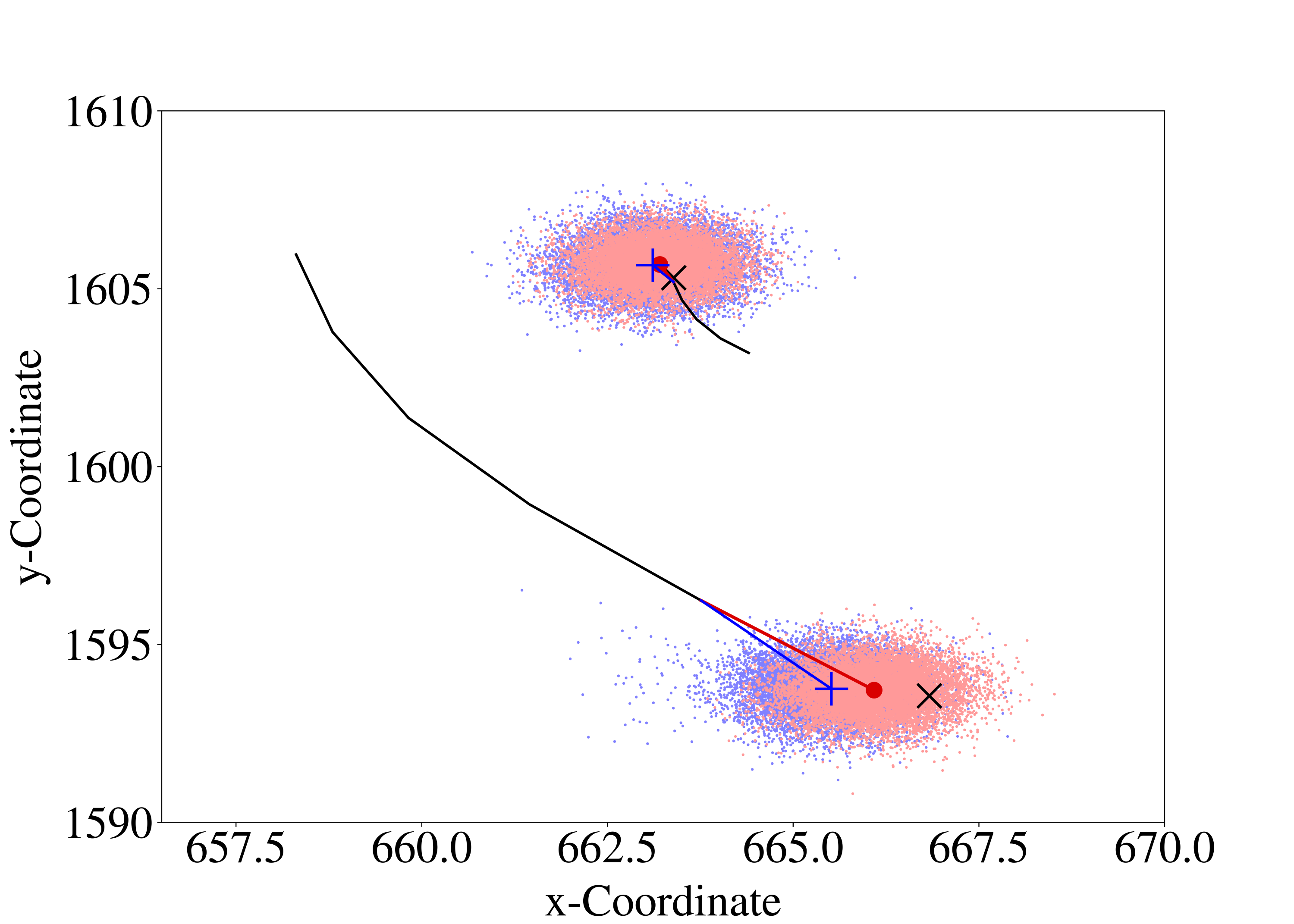}
	}
\caption{Visualization of the improved prediction performance related to the proposed neural-enhanced motion model. Prediction results for three different time steps are shown. In particular, predicted 2-D position information is indicated. Figs.~(a)--(c) show the prediction result for the pedestrians class, while Figs.~(d)--(f) show the prediction result for the car class. The model-based (MB) PDF is the predicted PDF obtained by the CV model and is represented by blue particles. The corresponding mean is indicated by a blue ``+". The neural enhanced (NE) PDF is the predicted PDF obtained by the neural network. The corresponding mean is marked by a red dot. The ground truth of position at each time step is marked by a black ``x''. For a fair prediction comparison, both the model-based and the neural-enhanced approach share the same track history. It can be seen that the neural-enhanced approach can provide prediction results that are closer to the ground truth\vspace{-3mm}. }
\label{fig:prediction}
\end{figure*}

\begin{table*}[!t]
	\centering
	\caption{Performance Comparison of Different Prediction Step Strategies}
	\begin{tabular}{l|c|c|c|c|c|c|c}
		\hline
		\multirow{2}{*}{\textbf{Method}} & \multicolumn{7}{c}{{\textbf{MSE} $\downarrow$}} \\
		\cline{2-8}
		& \textbf{bicycle} & \textbf{bus} & \textbf{car}& \textbf{motorcycle}& \textbf{pedestrian}& \textbf{trailer} & \textbf{truck} \\
		\hline
		CV Model (Baseline) & 0.187 & 2.572 & 0.434 & 0.450 & 0.090 & 0.462 & 0.469\\
		Object History (Proposed) & 0.170 & 2.090 & 0.299 & 0.430 & 0.061 & 0.407 & 0.365\\
		Object and Neighbors History (Proposed) & 0.160 & 1.900 & 0.285 & 0.405 & 0.058 & 0.400 & 0.356\\
		\hline
	\end{tabular}
	\label{tab:mse_pred}
		\vspace{-3mm}
\end{table*}

\begin{table}[t]
	\centering
	\caption{Performance Comparison of Different Prediction Step Implementations}
	\begin{tabular}{c|c|c|c}
		\hline
		{\textbf{Method}} & {{\textbf{AMOTA} $\uparrow$}} & {{\textbf{AMOTP} $\downarrow$}} & {{\textbf{FPS} $\uparrow$}} \\
		\hline
		No SPs & 76.1 & 51.6 & 4.6 \\
		Object SP Generation & 76.6 & 50.2 & 4.0\\
						Joint SP Generation & 76.6 & 50.0 & 2.0 \\
		\hline
	\end{tabular}
	\label{tab:ablation_pred_SP}
		\vspace{-3mm}
\end{table}
We also conduct ablation studies to evaluate the influence of the individual neural-enhancements for motion and measurement model. All results presented in this section are evaluated based on the nuScenes validation dataset and by making use of the FocalFormer \cite{YiZhiYu:C23-Detector-Focalformer3d} detector. Table \ref{tab:ablation_pred_update} presents the contributions of the prediction and update enhancements to the performance of the proposed method. The results demonstrate that the combination of both enhancements yields the best performance, achieving an AMOTA of 77.2 and an AMOTP of 49.7. When only applying one of the two enhancements, only a partial improvement is obtained. In what follows, we present ablation studies for the neural-enhanced motion and measurement models. In the ablation study of the motion model, the measurement model is purely model-based and vice versa.

\begin{figure*}[t]
	\centering
	\subfigure[Ground Truth]{
		\includegraphics[width=2.2in]{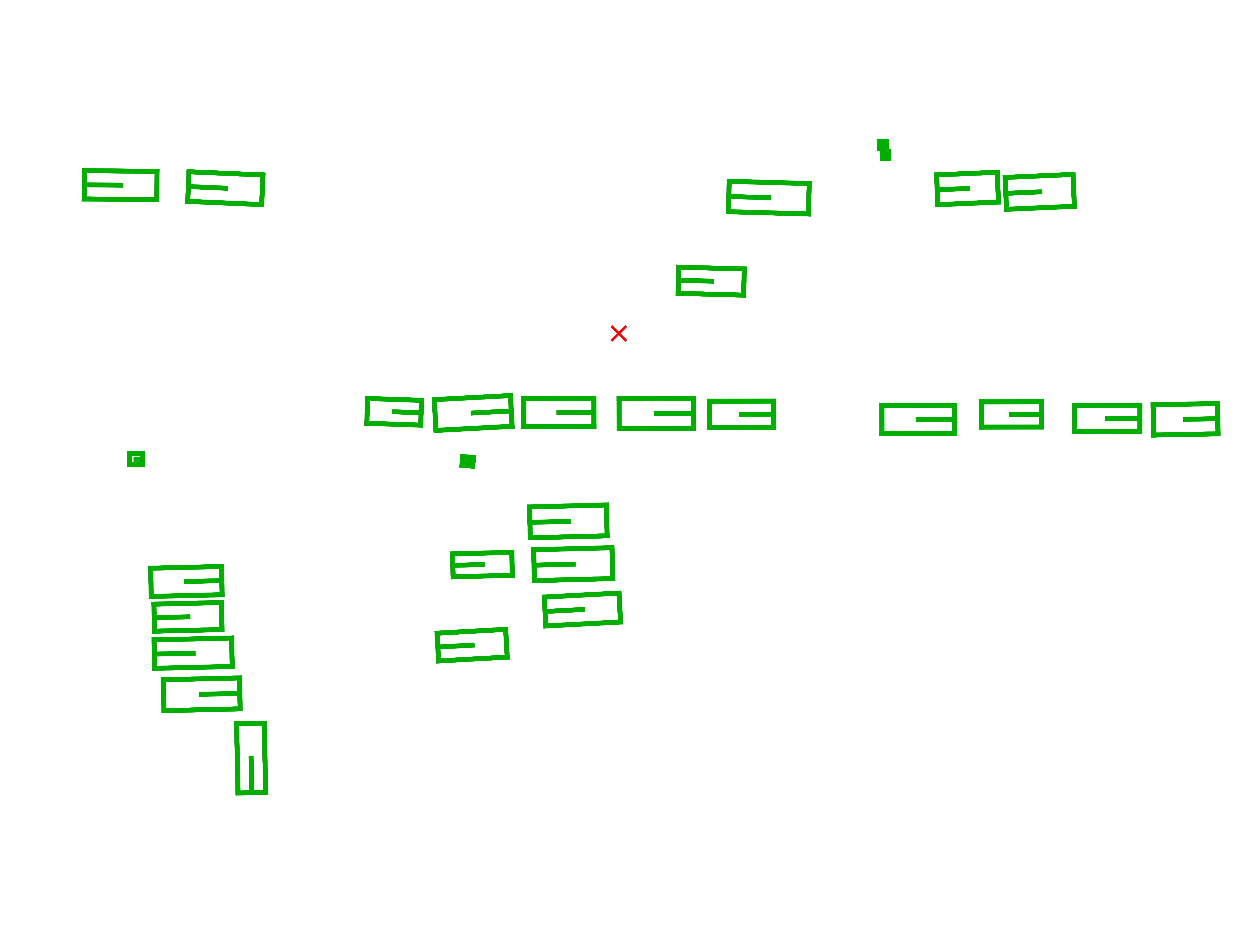}
	}
	\subfigure[Estimation Result for $\mathpzc{f}_{\mathrm{af}}(\checkmark)$ and $\mathpzc{f}_{\mathrm{fpr}}(\times)$]{
		\includegraphics[width=2.2in]{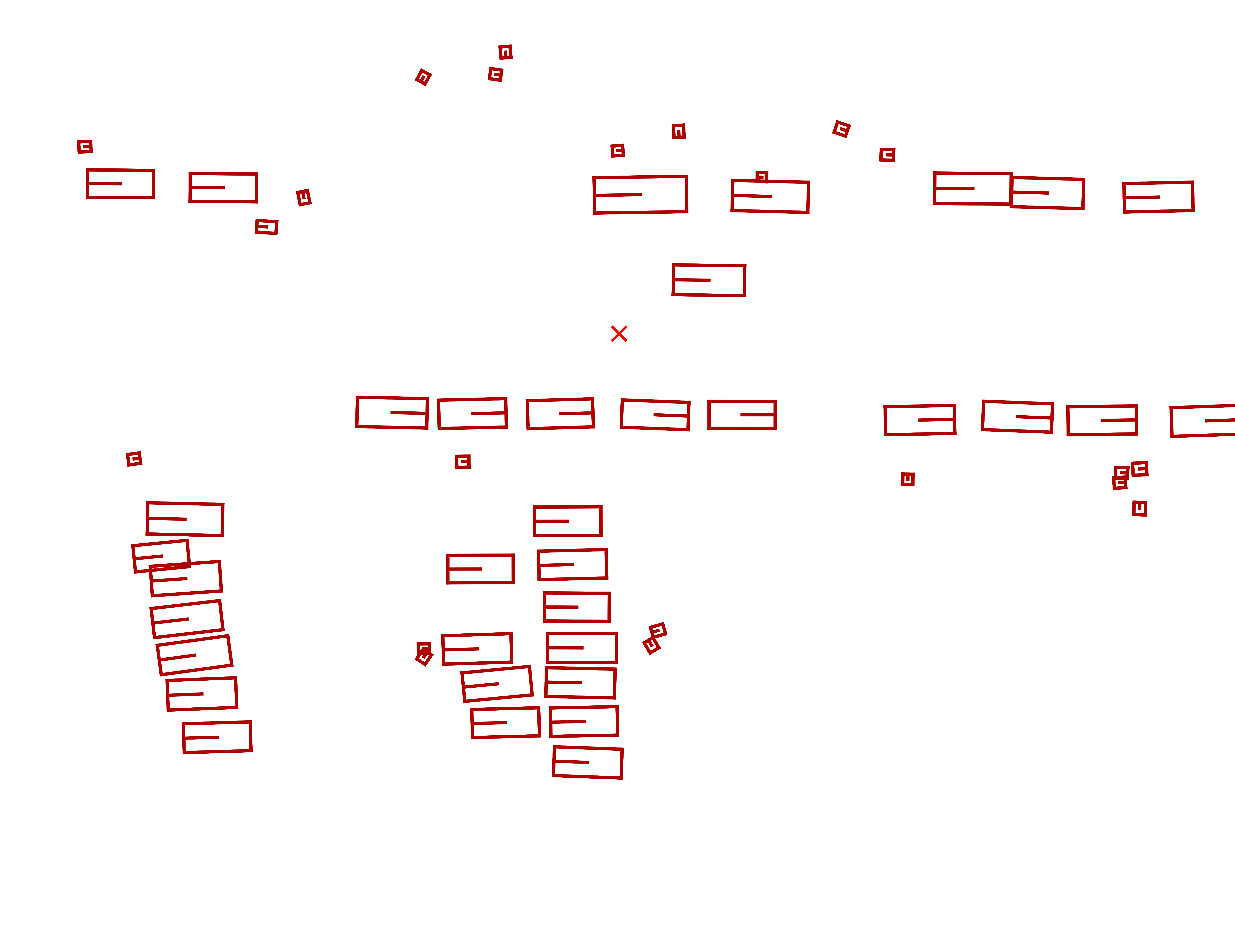}
	}
	\subfigure[Estimation Result for $\mathpzc{f}_{\mathrm{af}}(\checkmark)$ and $\mathpzc{f}_{\mathrm{fpr}}(\checkmark)$]{
		\includegraphics[width=2.2in]{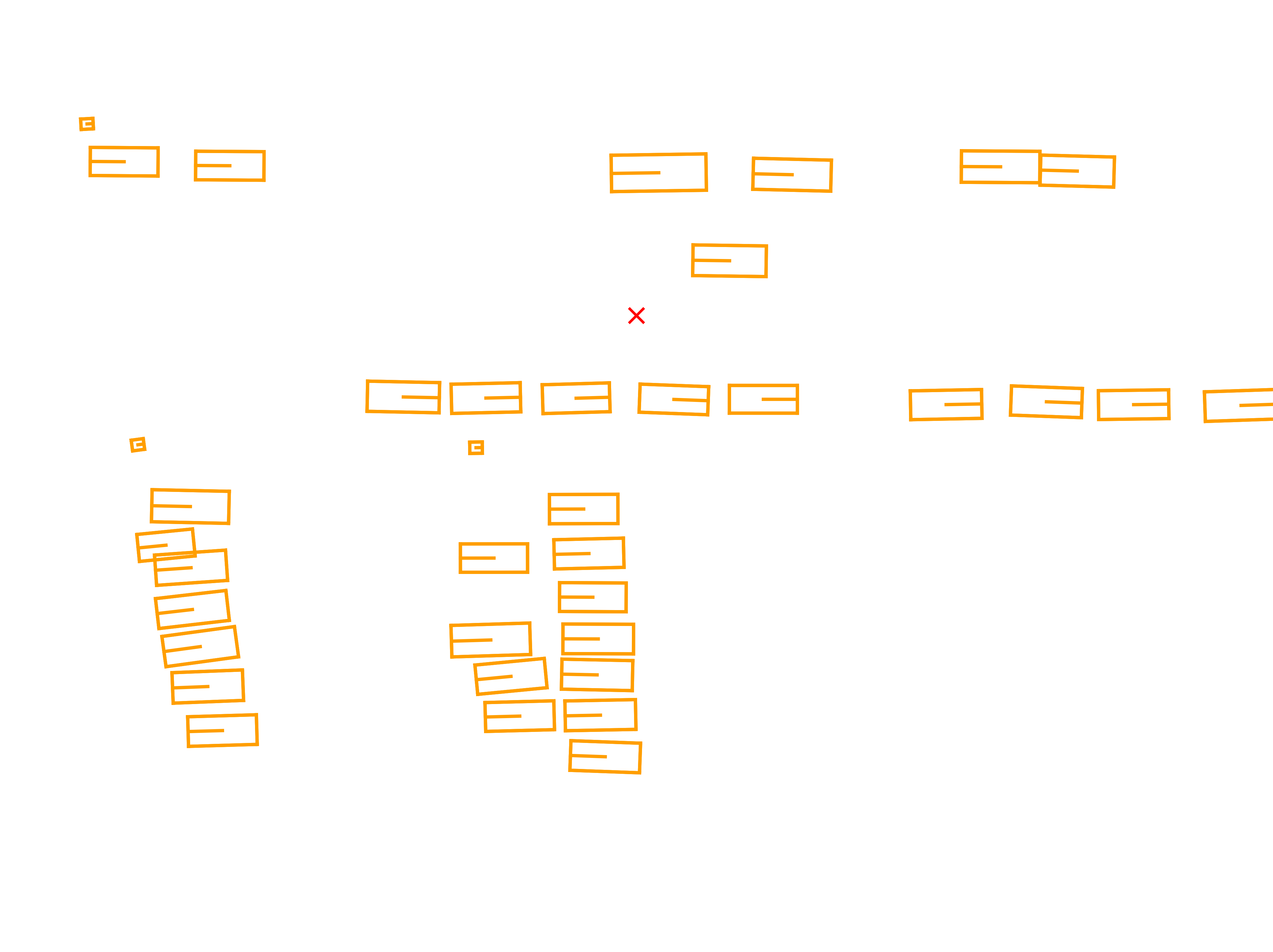}
	}
	\caption{Tracking results for cars and pedestrians classes at one time step. The ground truth in (a) is compared with the proposed neural-enhanced update step. The neural enhanced update step is performed without false positive rejection (b) and with false positive rejection factors (c). Object declaration is performed by setting $T_{\mathrm{dec}} \rmv=\rmv 0.7$. It can be seen that the result in (c) is closer to the result in (a) by having fewer false positives compared to (b).}
	\label{fig:false_alarm_factor}
\end{figure*}

\begin{figure*}[t]
	\centering
	\subfigure[Likelihood Function -- Three Cars and Clutter]{
		\includegraphics[width=3.4in]{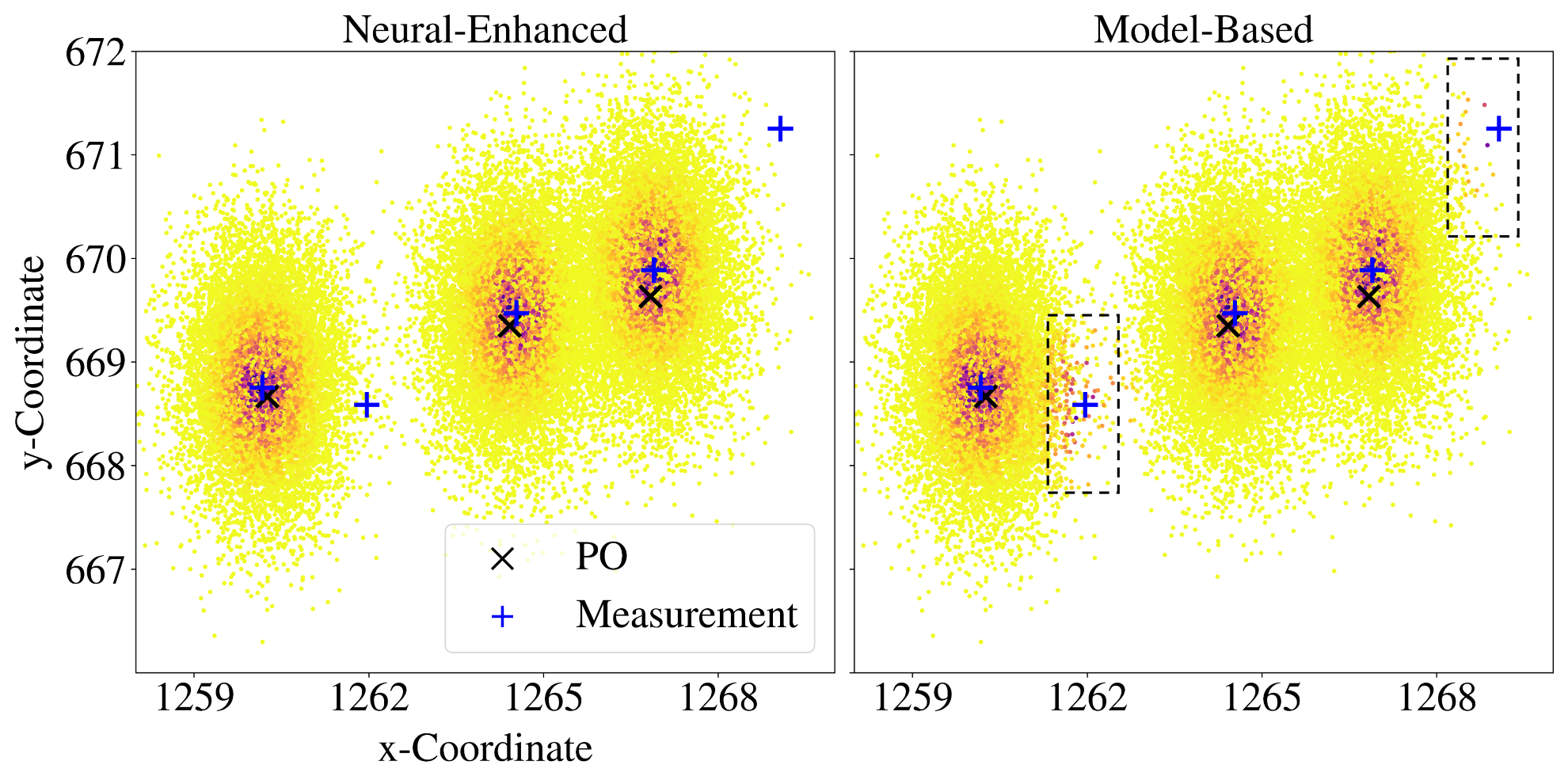}
	}
	\subfigure[Likelihood Function --  Two Close Pedestrians]{
		\includegraphics[width=3.4in]{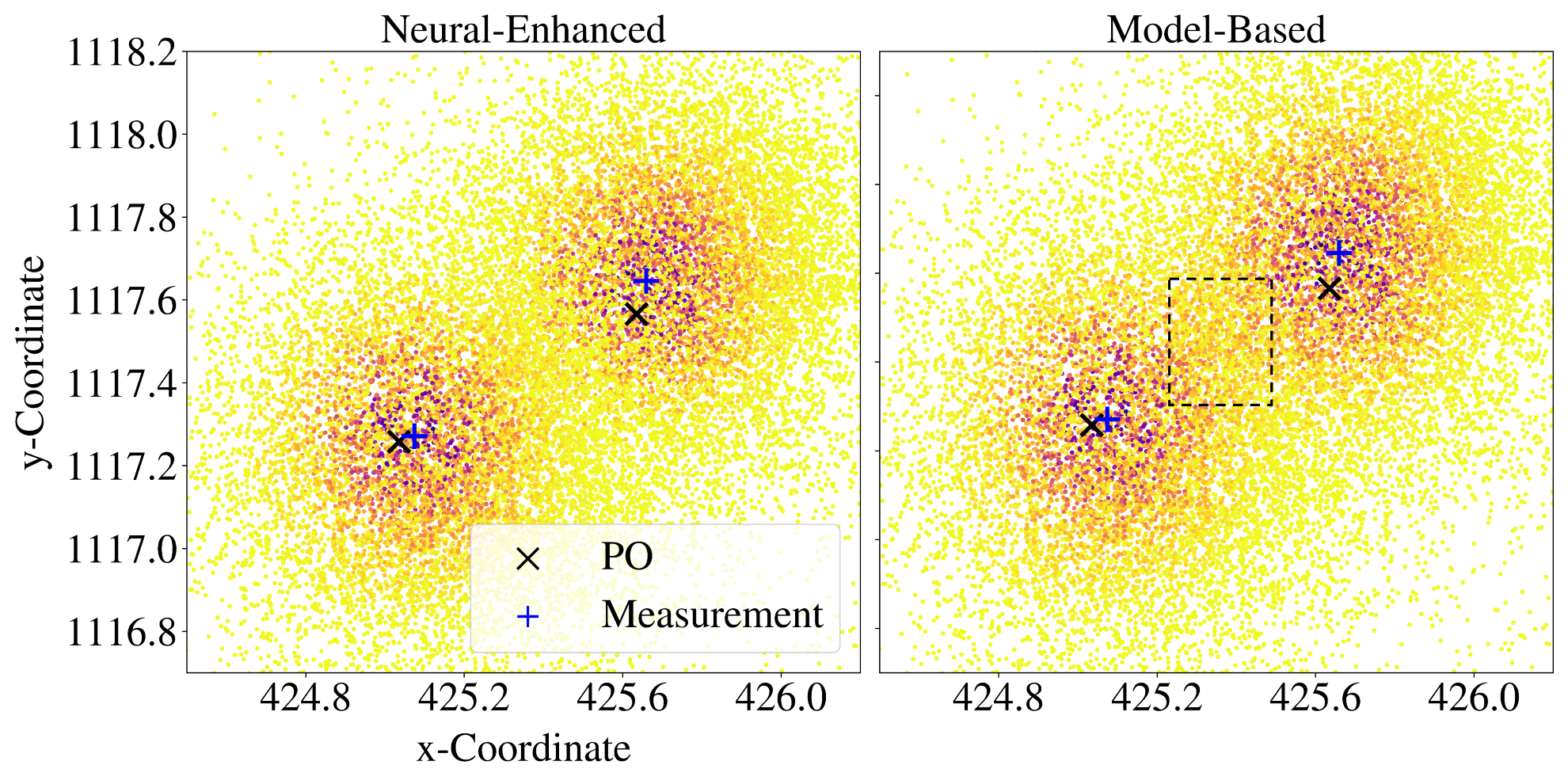}
	}
	\caption{Visualization of two examples that indicate how the affinity factors used by the neural-enhanced measurements update step can lead to improvements compared to a model-based measurement update step. Predicted particles and their corresponding weights provided by $l_{\mathrm{ne}}(\M{z}_k|\underline{\M{y}}^i_{k})$ and $l(\M{z}_k|\underline{\M{y}}^i_{k})$ are shown. A redder color indicates a higher particle weight. The black $\times$ denotes predicted position and the blue $+$ denotes measurements. For this visualization, the false positive rejection factor is set to one. The neural-enhanced measurements update step assigns smaller weights to false positive measurements in (a) and more clearly distinguish between the two pedestrians in (b).}
	\label{fig:afinity_factor}
\end{figure*}

\par First, we evaluate the contributions of historical tracks and object interactions to prediction accuracy. The evaluation is performed using noisy versions of ground truth tracks as input for one-step prediction. Noisy versions of ground truth tracks are obtained by matching the closest measurement to ground truth tracks using the Hungarian algorithm. The resulting sequence of measurements is used as a noisy version of the ground truth track. The results, computed for different object types, are quantified using the MSE of the position by comparing the prediction results with the ground truth track. Using the noisy versions of the ground truth track makes the prediction task more challenging and makes performance differences more pronounced. This study compares approaches with three different motion models: (i) a baseline method that relies on a Markovian constant velocity (CV) motion model, (ii) a simplified version of the proposed method with learnable non-Markovian motion model which relies solely on the history of the to-be-predicted object state but does not include states of neighboring object states, and (iii) the proposed 
method with the complete learnable non-Markovian motion model which also make use of the history of the to-be-predicted object state and neighboring object states. The results shown in Table \ref{tab:mse_pred} indicate the importance of using the history of the to-be-predicted object state and the history of neighboring states into account.

\par Second, we evaluate the proposed approach in tracking scenarios where performance is assessed using AMOTA and AMOTP metrics. This ablation study aims at comparing three processing strategies for the prediction step described in Section~\ref{sec:PredictionStep}. Strategy (i) only uses the sample mean for both to-be-predicted object state and neighboring object state as input for the joint transition function provided by the proposed neural network architecture; this strategy ignores the uncertainty quantification of both to-be-predicted and neighboring object states. Strategy (ii) generates  SPs only by making use of uncertainty quantification related to the to-be-predicted object state, and uses the sample mean of neighboring object states as input for the joint transition function; this strategy ignores the uncertainty quantification of neighboring object states. Strategy (iii) generates SPs by making use of uncertainty quantification for objects and their neighbors as discussed in Section \ref{sec:PredictionStep}.

The results, presented in Table \ref{tab:ablation_pred_SP}, demonstrate that propagating uncertainty quantification by means of SP can significantly improve prediction performance. However, joint SP generation and separate SP generation achieve quite similar performance. Table \ref{tab:ablation_pred_SP} also reports average Frames Per Second (FPS) to quantify the runtime of the three prediction strategies. FPS measurements have been obtained based on a laptop with a \textit{13th Gen Intel(R) Core(TM) i9-13900HX CPU} and an \textit{RTX 4060 GPU}. Note that the separate strategy (ii) results in a runtime that is roughly reduced by 50 \% compared to the joint strategy (iii). The visualization of the improved prediction performance related to the proposed neural-enhanced motion model is provided in Fig.~\ref{fig:prediction}. 

We also conducted an ablation study for the neural-enhanced measurement model. Here, we compare the performance of the neural-enhanced update step that uses both affinity factors (denoted $\mathpzc{f}_{\mathrm{af}}$) and false alarm rejection factors (denoted $\mathpzc{f}_{\mathrm{fpr}}$), with neural-enhanced update steps that only use affinity factors and false alarm positive factors, respectively. It can be seen in Table \ref{tab:ablation_factors} that the proposed approach yields the largest improvement in AMOTA performance only when both factors are considered. Both factors provide significant performance improvements. 
\par  Table \ref{tab:ablation_feature_for_ass} shows the contributions of each type of feature to factor $\mathpzc{f}_{\mathrm{af}}$. It can be seen that while only a subset of features already increases tracking performance compared to model-based MOT, the best result is achieved when the complete set of features is used. Another interesting insight is that even if only low-dimensional features are considered (i.e., position and velocity), this also leads to improved tracking performance. Notably, these low-dimensional features are already used within model-based MOT. This insight can be explained by the fact that the measurement model used by model-based MOT cannot accurately describe the underlying data-generating process, and only the proposed neural architecture can capture certain relationships of kinematic features with object states.

\par The positive impact of affinity factors is visualized in Fig.~\ref{fig:afinity_factor}. Here, false positive rejection factors are set to one to make sure their effects are omitted. In particular, predicted particles weighted by the functions $l_{\mathrm{ne}}(\M{z}_k|\underline{\M{y}}^i_{k})$ and $l(\M{z}_k|\underline{\M{y}}^i_{k})$ used for model-based and neural-enhanced measurement update are shown. Fig.~\ref{fig:afinity_factor}(a) depicts a scenario with three moving cars. Particles and their corresponding value of $l_{\mathrm{ne}}(\M{z}_k|\underline{\M{y}}^i_{k})$ and $l(\M{z}_k|\underline{\M{y}}^i_{k})$ for the three moving cars are shown. It can be observed that, contrary to the model-based measurement update, the neural-enhanced measurement update assigns lower particle weights to particles near false-positive measurements. Fig.~\ref{fig:afinity_factor}(b) illustrates another potential advantage of the neural-enhanced update step in a scenario with two pedestrians. In particular, it can be seen that the neural-enhanced update step makes it possible to more clearly distinguish between the two pedestrians by assigning smaller weights to particles located in the area between the two pedestrians. Fig.~\ref{fig:false_alarm_factor}, shows the positive effect of false positive rejection factors. In particular, it can be seen that the neural-enhanced update step with false positive rejection factors can clearly reduce the number of false positive object state estimates.

\begin{table}
	\centering
	\caption{Performance Evaluation -- Contributions of Likelihood Factors $\mathpzc{f}_{\mathrm{af}}$ and $\mathpzc{f}_{\mathrm{fpr}}$}
	\label{tab:ablation_factors}
	\begin{tabular}{c|c|c} 
		\hline
		\textbf{Method} &{\textbf{AMOTA}$\uparrow$} &\textbf{AMOTP}$\downarrow$\\ 
		\hline
		$\mathpzc{f}_{\mathrm{af}}(\times), \mathpzc{f}_{\mathrm{fpr}}(\times)$ &75.5&52.0\\   
		$\mathpzc{f}_{\mathrm{af}}(\times),  \mathpzc{f}_{\mathrm{fpr}}(\checkmark)$&75.9&50.5\\ 
        $\mathpzc{f}_{\mathrm{af}}(\checkmark),  \mathpzc{f}_{\mathrm{fpr}}(\times)$&76.4&51.0\\ 
		$\mathpzc{f}_{\mathrm{af}}(\checkmark),  \mathpzc{f}_{\mathrm{fpr}}(\checkmark)$ &76.7&49.5\\ 
		\hline
	\end{tabular}
\end{table}

\begin{table}
	\centering
	\caption{Performance Evaluation -- Contributions of Different Types of Features Used for $\mathpzc{f}_{\mathrm{af}}$ Computation}
	\label{tab:ablation_feature_for_ass}
	\begin{tabular}{c|c|c} 
		\hline
		\textbf{Method} &{\textbf{AMOTA}$\uparrow$} &\textbf{AMOTP}$\downarrow$\\ 
		\hline
		Model-Based &75.9&50.5\\   
		Kinematic Feature (Position and Velocity only) &76.2&51.5\\   
		Kinematic Feature &76.3&51.0\\ 
		Kinematic Feature and Shape Feature 1 &76.6&50.0\\ 
		Kinematic Feature and Shape Feature 1 and 2 &76.7&49.5\\ 
		\hline
	\end{tabular}
\end{table}

\section{Conclusion}
In this paper, we developed a hybrid model-based and data-driven method for multiobject tracking (MOT.) The key idea is to use neural networks to enhance specific aspects of the statistical model in model-based Bayesian MOT that have been identified as overly simplistic. The introduced neural-enhanced motion model captures the influence of historical object states and object interactions while the neural-enhanced measurement model makes use of complicated object shape information for data association and false positive measurement rejection. To obtain a good tradeoff between runtime and estimation accuracy, we use sigma points (SPs) in the neural-enhanced prediction step to evaluate the proposed learnable non-Markovian state motion model with object interactions efficiently, and particles for accurate neural-enhanced measurement update. Our numerical analysis demonstrates that the proposed method efficiently enhances the statistical model used for MOT and, applied to the publicly available nuScenes dataset, achieves state-of-the-art MOT performance. We believe that the proposed hybrid framework can serve as a guideline for the design of future MOT architectures. A potential direction for future research includes an application to underwater tracking problems.

\section*{Acknowledgment}
The authors thank Mr.~Xinshuang Liu and Prof.~Bhaskar D. Rao for illuminating discussions.

\begin{IEEEbiography}[{\includegraphics[width=1in,height=1.25in,clip,keepaspectratio]{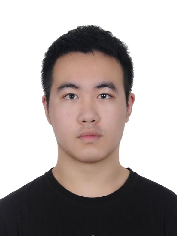}}]{Shaoxiu~Wei} (Student Member, IEEE) received the B.E. degree in communication engineering from the University of Electronic Science and Technology of China, Chengdu, China, in 2022. He is currently working toward the Ph.D. degree in statistical signal processing with the University of California San Diego, La Jolla, CA, USA. His research interests include machine learning, statistical signal processing, and wireless sensing.
\end{IEEEbiography}

\begin{IEEEbiography}[{\includegraphics[width=1in,height=1.25in,clip,keepaspectratio]{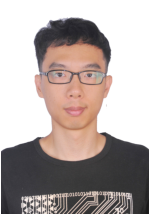}}]{Mingchao~Liang} (Student Member, IEEE) received the B.Eng. degree in electronic engineering from the Chinese University of Hong Kong, Hong Kong SAR, China, in 2018 and the Ph.D. degree in electrical and computer engineering from the University of California San Diego, La Jolla, CA, USA, in 2025. He is currently a Research Scientist at Meta Platforms in Menlo Park, CA, USA. Mr. Liang's research interests include localization and tracking algorithms, graphical models, machine learning, and deep learning. He was the recipient of a \textit{Qualcomm Innovation Fellowship} in 2022. Mingchao serves as a reviewer for the \textit{IEEE Transactions on Aerospace and Electronic Systems}, the \textit{IEEE Transactions on Wireless Communications}, and the \textit{IEEE Transactions on Signal Processing}.
\end{IEEEbiography}

\begin{IEEEbiography}[{\includegraphics[width=1in,height=1.25in,clip,keepaspectratio]{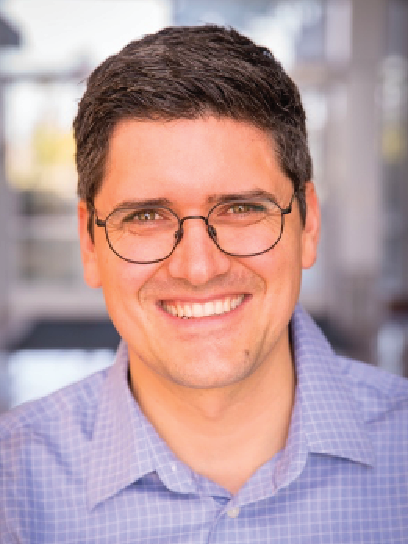}}]{Florian~Meyer} (Member, IEEE) received the M.Sc. and Ph.D. degrees (with highest honors) in electrical engineering from TU Wien, Vienna, Austria, in 2011 and 2015, respectively. He is an Associate Professor with the University of California San Diego, La Jolla, CA, USA, jointly appointed between the Scripps Institution of Oceanography and the Electrical and Computer Engineering Department. From 2017 to 2019, he was a Postdoctoral Fellow and Associate with the Laboratory for Information \& Decision Systems, Massachusetts Institute of Technology, Cambridge, MA, USA, and from 2016 to 2017, he was a Research Scientist with the NATO Centre for Maritime Research and Experimentation, La Spezia, Italy. His research interests include statistical signal processing, high-dimensional and nonlinear estimation, inference on graphs, inverse problems, multiobject tracking (MOT), and simultaneous localization and mapping (SLAM). He is the recipient of the 2021 ISIF Young Investigator Award, a 2022 NSF CAREER Award, a 2022 DARPA Young Faculty Award, and a 2023 ONR Young Investigator Award. He is currently an Associate Editor for \textit{IEEE Transactions on Signal Processing}. Dr. Meyer also served as an Associate Editor for the \textit{IEEE Transactions on Aerospace and Electronic Systems} from 2021 to 2023 and the \textit{ISIF Journal of Advances in Information Fusion} from 2019 to 2022.
\end{IEEEbiography}

\renewcommand{\baselinestretch}{1}
\bibliographystyle{IEEEtran}
\bibliography{IEEEabrv,references}

\end{document}